\documentclass[sigconf]{acmart}
\def\network{FMNet}
\usepackage{enumitem}
\usepackage{hyperref}
\hypersetup{
    colorlinks=true,
    linkcolor=red,
    urlcolor=cyan,
    }

\AtBeginDocument{%
  \providecommand\BibTeX{{%
    \normalfont B\kern-0.5em{\scshape i\kern-0.25em b}\kern-0.8em\TeX}}}

\copyrightyear{2022}
\acmYear{2022}
\setcopyright{acmcopyright}
\acmConference[MM '22] {Proceedings of the 30th ACM International Conference on Multimedia }{October 10--14, 2022}{Lisboa, Portugal.}
\acmBooktitle{Proceedings of the 30th ACM International Conference on Multimedia (MM '22), October 10--14, 2022, Lisboa, Portugal}
\acmPrice{15.00}
\acmISBN{978-1-4503-9203-7/22/10}
\acmDOI{10.1145/3503161.3547978}
\begin{document}
%\fancyhead{}
% \title{Less for Smooth: Exploiting Frame Masking for\\ Consistent Video Depth Estimation}
\title[Consistent Video Depth Estimation with Masked Frames Modeling]{Less is More: Consistent Video Depth Estimation with \\ Masked Frames Modeling}
%Less for Smooth: Frame Masking for Video Depth Estimation
\author{Yiran Wang}
\email{wangyiran@hust.edu.cn}
\affiliation{%
  \institution{School of AIA, Huazhong University of Science and Technology}
  \country{}
  %\streetaddress{P.O. Box 1212}
  %\city{Wuhan}
  %\state{Hubei}
  %\country{China}
}

\author{Zhiyu Pan}
\email{zhiyupan@hust.edu.cn}
\affiliation{%
  \institution{School of AIA, Huazhong University of Science and Technology}%Huazhong University of Science and Technology}
  \country{}
  %\streetaddress{P.O. Box 1212}
  %\city{Wuhan}
  %\state{Hubei}
  %\country{China}
}

\author{Xingyi Li}
\email{xingyi_li@hust.edu.cn}
\affiliation{%
  \institution{School of AIA, Huazhong University of Science and Technology}%Huazhong University of Science and Technology}
  \country{}
  %\streetaddress{1 Th{\o}rv{\"a}ld Circle}
  %\city{Wuhan}
  %\country{China}
}
%\email{larst@affiliation.org}

\author{Zhiguo Cao}
\email{zgcao@hust.edu.cn}
\affiliation{%
  \institution{School of AIA, Huazhong University of Science and Technology}%Huazhong University of Science and Technology}
  %\city{Wuhan}
  \country{}
}

\author{Ke Xian}
\email{kexian@hust.edu.cn}
\affiliation{%
 \institution{School of AIA, Huazhong University of Science and Technology}%Huazhong University of Science and Technology}
 \country{}
}
\authornote{Corresponding author.}

\author{Jianming Zhang}
\email{jianmzha@adobe.com}
\affiliation{
  \institution{Adobe Research}
  \country{}
}

%\renewcommand{\shortauthors}{Trovato and Tobin, \textit{et al.}}
\renewcommand{\shortauthors}{Yiran Wang et al.}
\begin{abstract}
Temporal consistency is the key challenge of video depth estimation. Previous works are based on additional optical flow or camera poses, which is time-consuming. By contrast, we derive consistency with less information. Since videos inherently exist with heavy temporal redundancy, a missing frame could be recovered from neighboring ones. Inspired by this, we propose the frame masking network (FMNet), a spatial-temporal transformer network predicting the depth of masked frames based on their neighboring frames. By reconstructing masked temporal features, the FMNet can learn intrinsic inter-frame correlations, which leads to consistency. Compared with prior arts, experimental results demonstrate that our approach achieves comparable spatial accuracy and higher temporal consistency without any additional information. Our work provides a new perspective on consistent video depth estimation. Our official project page is \href{https://github.com/RaymondWang987/FMNet}{\textcolor{magenta}{https://github.com/RaymondWang987/FMNet}}.

\vspace{-8pt}

\end{abstract}


\begin{CCSXML}
<ccs2012>
   <concept>
       <concept_id>10010147.10010178.10010224.10010225</concept_id>
       <concept_desc>Computing methodologies~Computer vision tasks</concept_desc>
       <concept_significance>500</concept_significance>
       </concept>
 </ccs2012>
 \vspace{-8pt}
\end{CCSXML}



%  \includegraphics[width=\textwidth]{fig1_v3.pdf}

\maketitle
\begin{figure}[!h]
    \centering
    \includegraphics[scale=0.30,trim=25 60 25 10,clip]{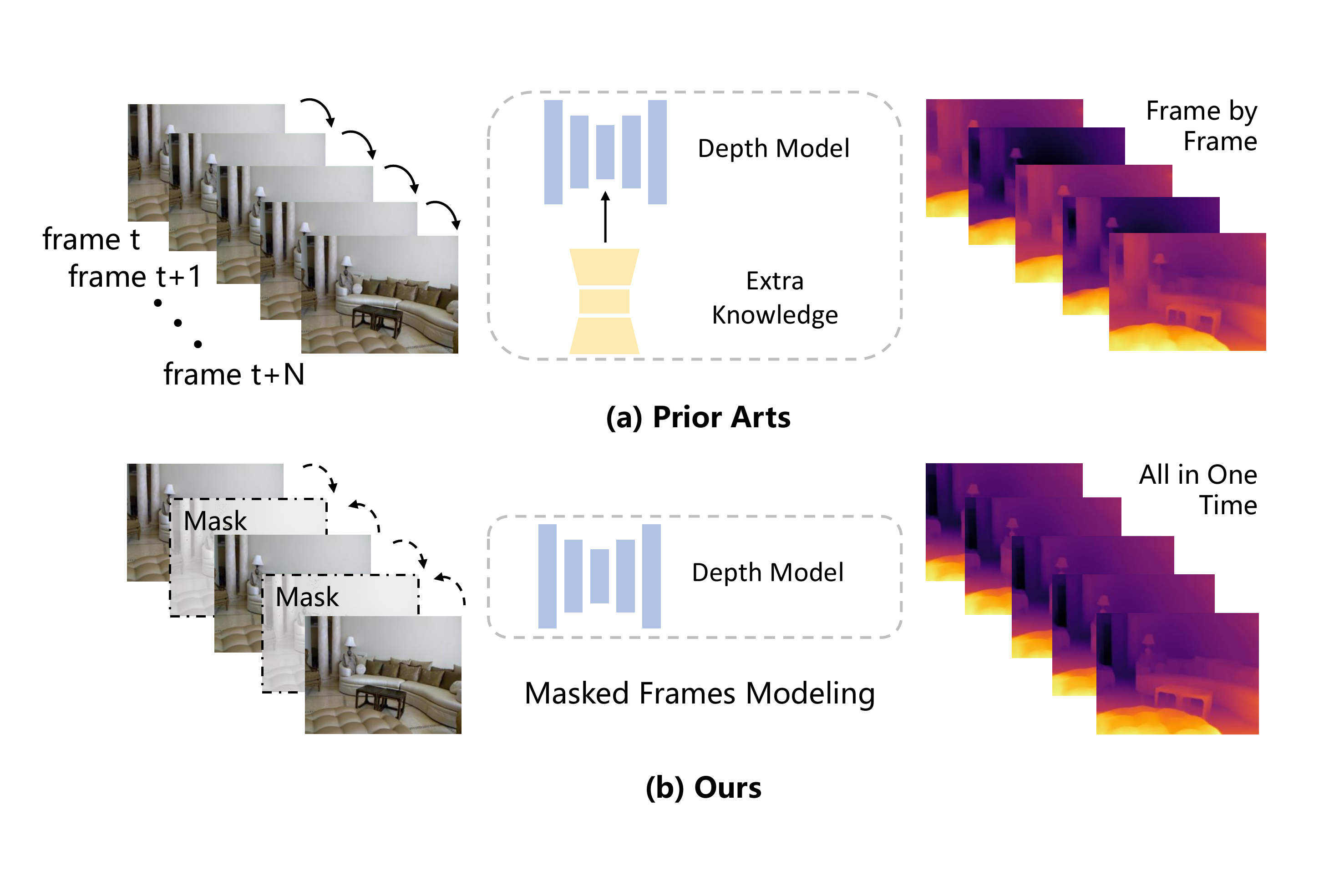}
    \caption{Prior arts leverage extra knowledge from optical flow~\cite{MM21,cvd,rcvd} or GANs~\cite{ST-CLSTM}. Most of them~\cite{ST-CLSTM,rcvd,MM21} transfer temporal knowledge serially and produce depth results frame by frame. By contrast, we derive the consistency with less information by masked frames modeling. %Previous methods~\cite{MM21,ST-CLSTM,rcvd} transfer the temporal knowledge serially and the depth results are produced frame by frame.
    Our method can produce depth results of all input frames in one time.}
    \label{fig:fig1}
\end{figure}
\vspace{-8pt}
\section{Introduction}
% Monocular video depth estimation~\cite{ST-CLSTM,MM21,cvd,rcvd,dycvd,deepv2d} requires both spatial accuracy and temporal consistency. In recent years, the spatial accuracy of single image depth estimation~\cite{midas,jrtz,dpt,monodepth2,kexian2020} has been significantly improved. However, the temporal consistency remains an open question. Some previous works~\cite{ST-CLSTM} mine temporal consistency in a generative and adversarial paradigm~\cite{gan}. However, when partitioning a video clip to several input sequences, their method would cause obvious flickering between sequences due to the seriality of their model.% \kexian{Be more specific, like what kind of limitations.}
% some other works~\cite{cvd,rcvd,MM21} exploit the consistency with the help of optical flow~\cite{flownet2,raft}, but their methods could be time-consuming. Those methods could fail when the additional information. Under this circumstance, we pursue a simple solution where the temporal consistency can be mined solely from the relation between original input frames.
Monocular video depth estimation plays a vital role in many applications, such as 2D-to-3D video conversion~\cite{n1}, scene reconstruction~\cite{deepv2d}, and bokeh rendering~\cite{videobokeh,bokehme}. To obtain pleasant and immersive experience, both spatial accuracy and temporal consistency are required. In recent years, the spatial accuracy of single image depth estimation~\cite{midas,dpt,kexian2020,DABC} has been significantly improved. However, the temporal consistency remains an open question. When deploying single image depth estimation approaches on videos, it would cause an obvious flickering problem due to the independent computation of each frame. To estimate consistent video depth, as shown in Fig.~\ref{fig:fig1}(a), most previous approaches~\cite{ST-CLSTM,MM21,cvd,rcvd,dycvd} model inter-frame correlations based on extra temporal clues. Such clues can be provided by additional computations, \textit{e.g.}, optical flow~\cite{raft,flownet2,MM21}, pose estimation~\cite{colmapsfm,cvd,rcvd}, and generative adversarial networks (GANs)~\cite{gan,ST-CLSTM}. As a consequence, those methods
totally fail when the temporal clues are inaccurate. Meanwhile, these approaches are time-consuming. For example, for a video of $244$ frames, CVD~\cite{cvd} takes about $40$ minutes for test-time training on $4$ NVIDIA Tesla M40 GPUs. This begs the question -- \textit{is it possible to achieve temporal consistency without explicitly modeling inter-frame correlations?}

%Inspired by the recent works~\cite{bert,beit,mae}, which use masked autoencoders to reconstruct the masked language fragments and image patches, we propose a frame masking network (FMNet) to estimate consistent video depth.
Languages, images, and videos are of high redundancy in nature. For example, we can still guess what a sentence means if some words are blocked; we can still recognize what a picture depicts if some patches are missing; and we can still imagine how an object moves if some video frames are removed. Inspired by this, we explore temporal consistency with implicit constraints by reconstructing the masked frames, which reduces the dependency on additional optical flow or camera poses. In particular, we design a subtask to recover depth maps of the masked frames based on their neighboring ones. As illustrated in Fig.~\ref{fig:fig1}(b), different from previous works, we introduce the masking and predicting paradigm to video depth estimation. Randomly masking some input frames, our model is trained to reconstruct the depth structures of unmasked frames. This can force our model to predict the depth results based on possibly relevant frames. In this manner, the inter-frame correlations are enhanced and the model acquires a larger temporal receptive field. Therefore, the video depth results can be more consistent.

We propose the frame masking network (FMNet) for consistent video depth estimation. The input video sequences are processed by a spatial feature extractor that generates the representation of each frame. To model the global inter-frame correlations, we adopt a transformer~\cite{transformer} architecture as the temporal feature extractor. However, we can only represent frames as feature maps to keep their spatial structures, which makes the cost of attention computation unacceptable. Inspired by ConvTransformer~\cite{ctrans}, we choose to approximate the attention computation of feature maps by convolution~\cite{lenet}. The temporal encoder randomly masks a certain portion of input frames and encodes the temporal correlation features based on the remaining frames. %\kexian{rephrase}The temporal correlations feature maps are completed by a learnable mask token to the feature maps sequence.
We use a learnable mask token to fill in the masked positions. The completed sequence is then processed by the temporal decoder and depth predictor to estimate depth structures of both masked and unmasked frames. In this manner, the temporal encoder can learn temporal inter-frame correlations implicitly. The temporal correlations knowledge enables the FMNet to generate consistent depth results.
% The first difference is the representation of basic elements, \textit{i.e.}, language words and image patches. Language words and image patches are represented by vectors through word2vector~\cite{w2v1,w2v2,w2v3} or flatten operation~\cite{VIT}. For the video depth estimation task, the basic elements are frames. To keep the spatial structure of frames, we can only represent them as feature maps rather than vectors. This leads to the other difference which is the relation encoding between basic elements. Prior arts~\cite{bert,gpt1,VIT,mae} employ transformer architecture~\cite{transformer} to process the vector embedding of each elements. When our representations for frames are feature maps, the cost of attention computation in the transformer is unacceptable. Inspired by ConvTransformer~\cite{ctrans}, we choose to approximate the attention computation of feature maps by convolution~\cite{lenet,alexnet,resnet}.

Experimental results demonstrate that our FMNet can generate depth results with both spatial accuracy and temporal consistency. Compared with other state-of-the-art approaches, quantitative and qualitative results show the superiority of our FMNet in temporal consistency. We also achieve comparable spatial accuracy with prior arts. Further analysis on the NYU Depth V2 dataset shows that a very high masking ratio benefits both accuracy and consistency. Different from previous works relying on optical flow, pose estimation, or GANs, our work provides a novel information-reductive perspective for temporal consistency in video depth estimation. The main contributions of this paper can be summarized as follows:
%, where only two frames remain in a clip,
\begin{itemize}[leftmargin=*]
    %kx
    \item We propose a masked video transformer for consistent video depth estimation without relying on optical flow, pose estimation, and GANs.
    %\item We design a masked frames predicting strategy for temporal consistency without optical flow, camera poses or GANs.
    \item To the best of our knowledge, we are the first to introduce the ConvTransformer for video depth estimation, which can encode inter-frame correlations in parallel.
    % \item We firstly introduce the ConvTransformer to video depth estimation, which can encode inter-frame relation in parallel.
    % \item Experimental results demonstate that our \network{} could achieve better temporal consistency without relying on optical flow, camera pose or GANs.
\end{itemize}

\section{RELATED WORK}
\subsection{Consistent Video Depth Estimation}
Consistent video depth estimation focuses on removing flickering in video depth results. It takes both spatial accuracy and temporal consistency into account. Unfortunately, approaches~\cite{kexian2018,kexian2020,midas,dpt,adabins,bts} for single image depth estimation fail to handle the temporal consistency. They can not model the inter-frame correlations, causing obvious flickering in video depth results. Various approaches are devoted to tackling this problem
based on two paradigms: test-time optimization and training-time modeling.

\textbf{Test-time optimization approaches} refine video depth results from existing models. CVD~\cite{cvd} refines depth models~\cite{midas,monodepth2} by warping frame pairs using camera poses~\cite{colmapsfm,colmapmvs} and optical flow~\cite{flownet2}. %Their method is sensitive to dynamic scenes. To solve this problem, Robust-CVD~\cite{rcvd} jointly optimizes camera poses and depth maps. Dynamic-video-depth~\cite{dycvd} achieves consistency by predicting the scene flow. However, these approaches collapse when optical flow and poses are inaccurate.
Their main limitation is that accurate camera poses are required. Unfortunately, pose estimation itself is a challenging problem and would fail in the presence of dynamic object motion. To address this, Robust-CVD~\cite{rcvd} jointly optimizes camera poses and depth alignments. Although their method improves the robustness of video depth estimation, such joint optimization is inefficient and time-consuming, as we explained earlier.

\textbf{Training-time modeling approaches} generate consistent video depth without the time-consuming test-time training process. ST-CLSTM~\cite{ST-CLSTM} applies LSTM and GAN for the temporal consistency. Cao \textit{et al.}~\cite{MM21} propose a convolutional spatial-temporal propagation network~\cite{spn} trained with knowledge distillation~\cite{kd0,kd1,kd2} and optical flow~\cite{flownet2}. These methods suffer from heavy training burdens with additional optical flow or GANs for supervision. By contrast, our work utilizes the masked frames predicting strategy and does not require any additional information.
\begin{figure*}
    \centering
    \includegraphics[scale=0.45,trim=0 20 0 0,clip]{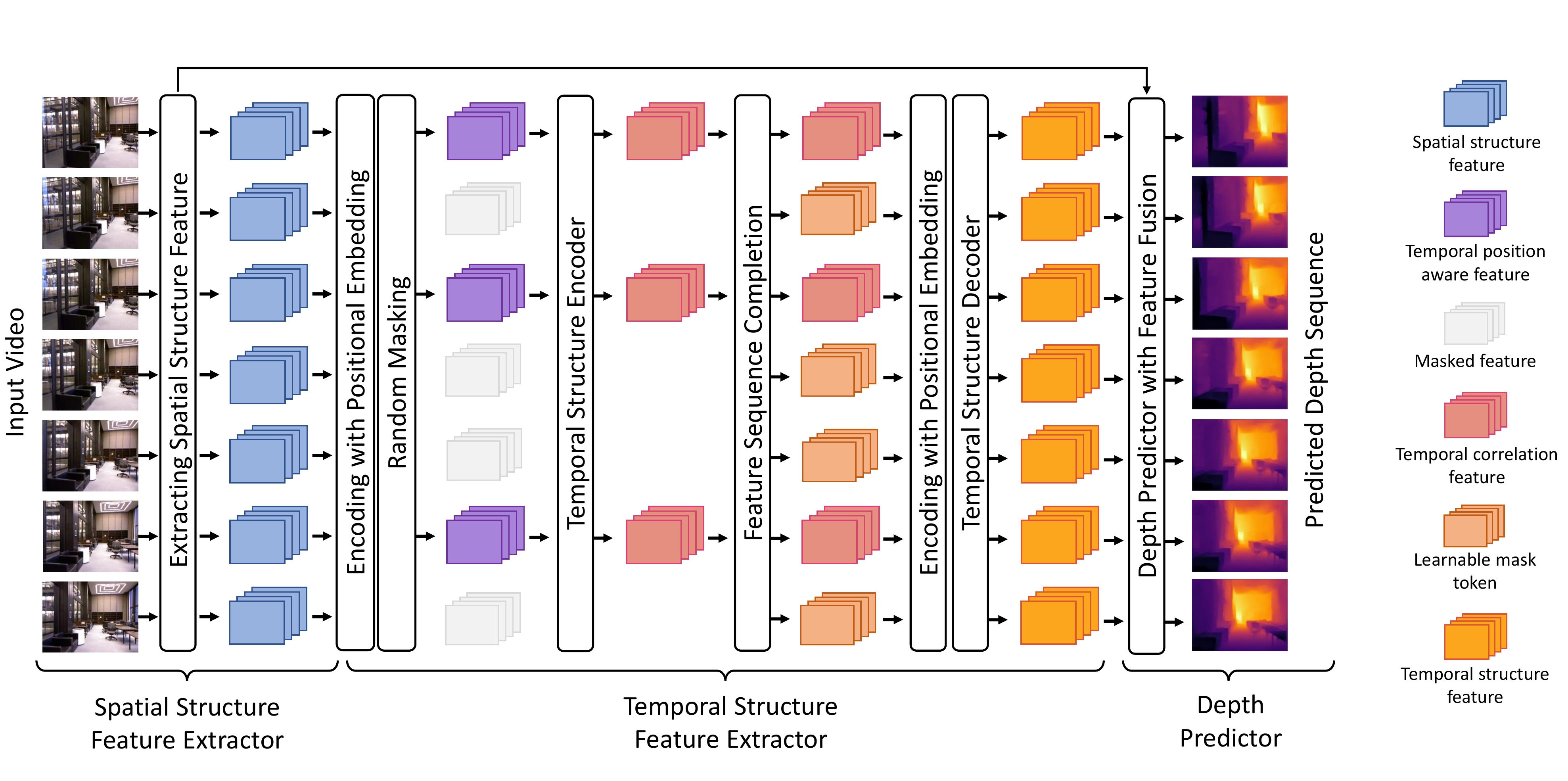}
\caption{Illustration of our FMNet for consistent video depth estimation. Taking $\textbf{N}$ consecutive frames as input, the spatial feature extractor generates the feature of each frame. The temporal structure encoder randomly masks a portion of input frames and encodes temporal correlation features. We use a learnable mask token to fill in the masked positions. The completed sequence is processed by the temporal structure decoder to restore the masked temporal structure features. Finally, the depth predictor fuses the spatial and temporal structure features to predict consistent depth maps of all input frames in one time.}
\label{fig:pipeline}
\end{figure*}
\subsection{Transformer}
RNNs and LSTMs~\cite{sxrnn,lstm} have dominated the sequence processing area for a long time until the emergence of transformer~\cite{transformer}. The transformer can model correlations among sequence elements in parallel, which can encode more global representations. Some works~\cite{VIT,swin,mvit,detr,newdetr} have shown the superiority of transformer in image-based tasks. For video data, ViViT~\cite{vivit} designs their model by factorizing the spatial and temporal dimension as a straightforward extension of ViT~\cite{VIT}. Liu \textit{et al.}~\cite{ctrans} replace the traditional attention mechanism with convolutional self-attention and introduce the ConvTransformer architecture. Their model can directly process video sequences, which is more appropriate for video-related tasks. In this work, we follow the idea of ConvTransformer~\cite{ctrans} to model inter-frame correlations.

\subsection{Masked Data Modeling}
%Approaches that share the idea of masked data predicting learn data representations by reconstructing the masked information.
Recent works explore to learn structure information by reconstructing the masked signals. For language data, GPT~\cite{gpt1,gpt2,gpt3} and BERT~\cite{bert} propose to predict the masked words in sentences, which shows superiority in natural language processing. For image data, iGPT~\cite{igpt} and BEiT~\cite{beit} convert the image patches to visual tokens and predict the masked ones. MAE~\cite{mae} directly masks the image patches and learns image representations by reconstructing the original image. To the best of our knowledge, we are the first to adopt the idea of masked data modeling to deal with the task of video depth estimation. We find that this strategy can also benefit the learning of consistent representations for videos.

\subsection{Structure from Motion}
Structure-from-motion (SFM) methods~\cite{banet,deepv2d} predict depth maps by feature matching over multiple frames. BA-Net~\cite{banet} regresses depth maps via feature-metric bundle adjustment. DeepV2D~\cite{deepv2d} alternately updates depth and camera motion. They conduct cost volumes and feature matching in their depth module. Those SFM-based methods highly rely on accurate camera poses. They only benefit static scenes but do not account for dynamical objects due to the failure of pose estimation. By contrast, our \network{} is not limited by pose estimation. Our method is also significantly faster than DeepV2D~\cite{deepv2d} because the feature matching and pose estimation are inefficient and time-consuming.

\section{PROPOSED METHOD}
\subsection{Overview}
Here we present an overview of our method. Fig.~\ref{fig:pipeline} shows the technical pipeline of our \network{} framework. Given $N$ consecutive frames as input, the spatial structure feature extractor, \textit{i.e.}, a CNN encoder, extracts spatial features of input frames independently. It cannot model input-frame correlations. Although RNNs~\cite{sxrnn} and LSTMs~\cite{lstm} can somehow deal with the temporal correlations, we observe that ST-CLSTM~\cite{ST-CLSTM} generates obvious flickering (see Sec.~\ref{sec:mr}) between adjacent frames due to its seriality and locality. Hence, we naturally resort to transformer~\cite{transformer} for its parallelism and globality. Nevertheless, the original transformer is not appropriate for processing sequences of high-dimensional features due to the computational overhead of attention mechanism. Following~\cite{ctrans}, we adopt the idea of ConvTransformer, which can directly process features sequences without partitioning patches and flatten operation. To represent the chronological order of input frames, we add positional embeddings to each sequence of spatial features.

Recent MAE~\cite{mae} observes and utilizes the spatial redundancy of single images.
% By masking image patches and reconstructing them,
By masking and reconstructing image patches,
the MAE encoder~\cite{VIT} is forced to learn high-level semantic information. Inspired by this, our \network{} leverages the high redundancy of videos in temporal dimension. We design our masked frames predicting strategy to force the temporal structure feature extractor to learn temporal correlations among frames. To be specific, as illustrated in Fig.~\ref{fig:pipeline}, we randomly mask a portion of the spatial feature maps with positional embeddings. The remaining unmasked features are fed into our temporal structure encoder to build inter-frames temporal correlations. The outputs of the temporal structure encoder are feature maps corresponding to the unmasked frames with the same shape as its input. The next step is to
restore temporal structure features of the masked frames based on the remaining ones. We use a shared and learnable mask token to represent the masked positions and complete the full sequence of feature maps. We add positional embeddings to the full sequence again and feed it into our temporal structure decoder. Our temporal structure decoder has the same ConvTransformer architecture as the encoder, which has only one transformer layer. The heavy computational burden
of the full features sequence is only sustained by the lightweight one-layer decoder, while the six-layer encoder only takes the unmasked features as input. In this way, the computational cost of our method could be reduced compared with other methods based on optical flow or camera poses. The temporal structure decoder predicts the sequence of temporal structure features for all input frames.

Finally, we adopt our depth predictor, \textit{i.e.}, a CNN decoder, to restore depth maps of the input sequence. We employ five up-projection modules to improve the spatial resolution and decrease the channel numbers. We use the feature fusion module (FFM)~\cite{FFM1,FFM2} and skip connection from the spatial structure feature extractor to the depth predictor to fuse the spatial and temporal structure features. As for training, we adopt the widely-used scale-invariant loss~\cite{silog}. Our method does not need extra temporal loss functions based on camera poses, optical flow, or GANs. With the same training paradigm as single image depth estimation, our \network{} can predict video depth results with both spatial accuracy and temporal consistency because of our masked frames predicting strategy.

We will illustrate our ConvTransformer architecture in Sec.~\ref{sec:ctrans} and our masked frames predicting strategy in Sec.~\ref{sec:fm}. The loss function for training is in Sec.~\ref{sec:de}.
%In our case, $h = H/32$, $w = W/32$, and $c = 2048.$
\subsection{ConvTransformer}
\label{sec:ctrans}
Given a sequence of $N$ input video frames
$F = \{F_0,F_1,\cdots,F_{N-1} \}$ where $F_{i} \in \mathbb{R}^{H\times W\times 3}$, %$H, W$ denote frames height and width, respectively.
our ConvTransformer is built to model temporal correlations among input frames. However, taking the original video frames as direct input is inappropriate due to the high resolution. We consider our spatial structure feature extractor as feature embeddings to extract the spatial feature maps of input frames $f = \{f_0,f_1,\cdots,f_{N-1} \}$, where $f_{i} \in \mathbb{R}^{h\times w\times c}$, $h,w,$ and $c$ denote feature maps height, width, and the number of channels, respectively. If we denote the spatial structure feature extractor by $\mathcal{F}_{\theta_
{\mathcal{F}}} $, where $\theta_
{\mathcal{F}}$ refers to the parameters of $\mathcal{F}_{\theta_
{\mathcal{F}}} $, we extract the spatial feature maps of input frames using the shared extractor with same parameters:
\begin{equation}
    \label{equ:(1)}
    f_i = \mathcal{F}_{\theta_{\mathcal{F}}}(F_i), \quad i\in[0,N-1]\,.
\end{equation}

We use our spatial structure feature extractor to extract the spatial features and our ConvTransformer to represent the temporal correlation features among input frames. In this section, we do not consider our masked frames predicting strategy for the time being. In order to illustrate the ConvTransformer architecture, we assume the sequence of feature maps $f = \{f_0,f_1,\cdots,f_{N-1} \}$ as its input.

Similar to the original transformer~\cite{transformer}, the ConvTransformer needs positional embeddings to represent the chronological order of input feature maps. We adopt the same positional embeddings as~\cite{ctrans}, which are 3D tensors with the same shape as input feature maps. It can be expressed as an extension of the original transformer~\cite{transformer} sine and cosine positional embeddings:
\begin{equation}
    \begin{gathered}
    PE(pos,(x,y),2k) = \sin(pos/10000^{2k/c})\,, \\
    PE(pos,(x,y),2k+1) = \cos(pos/10000^{2k/c})\,, \label{equ:pos}
    \end{gathered}
\end{equation}
where $pos$ denotes the position of a certain feature map in the full sequence, $(x,y)$ denotes spatial locations in feature maps, $2k$ refers to the channel dimension, and $c$ denotes the feature maps channels amount. Given a feature map $f_i,i\in[0,N-1]$, its positional embedding $PE_i$ has the same shape $h\times w\times c$ as $f_i$. The positional embeddings can be directly added to the original feature maps:
\begin{equation}
    \label{equ:(3)}
    p_i = f_i + PE_i, \quad i\in [0,N-1]\,.
\end{equation}

\begin{figure}
    \centering
    \includegraphics[scale=0.34,trim=0 30 0 30,clip]{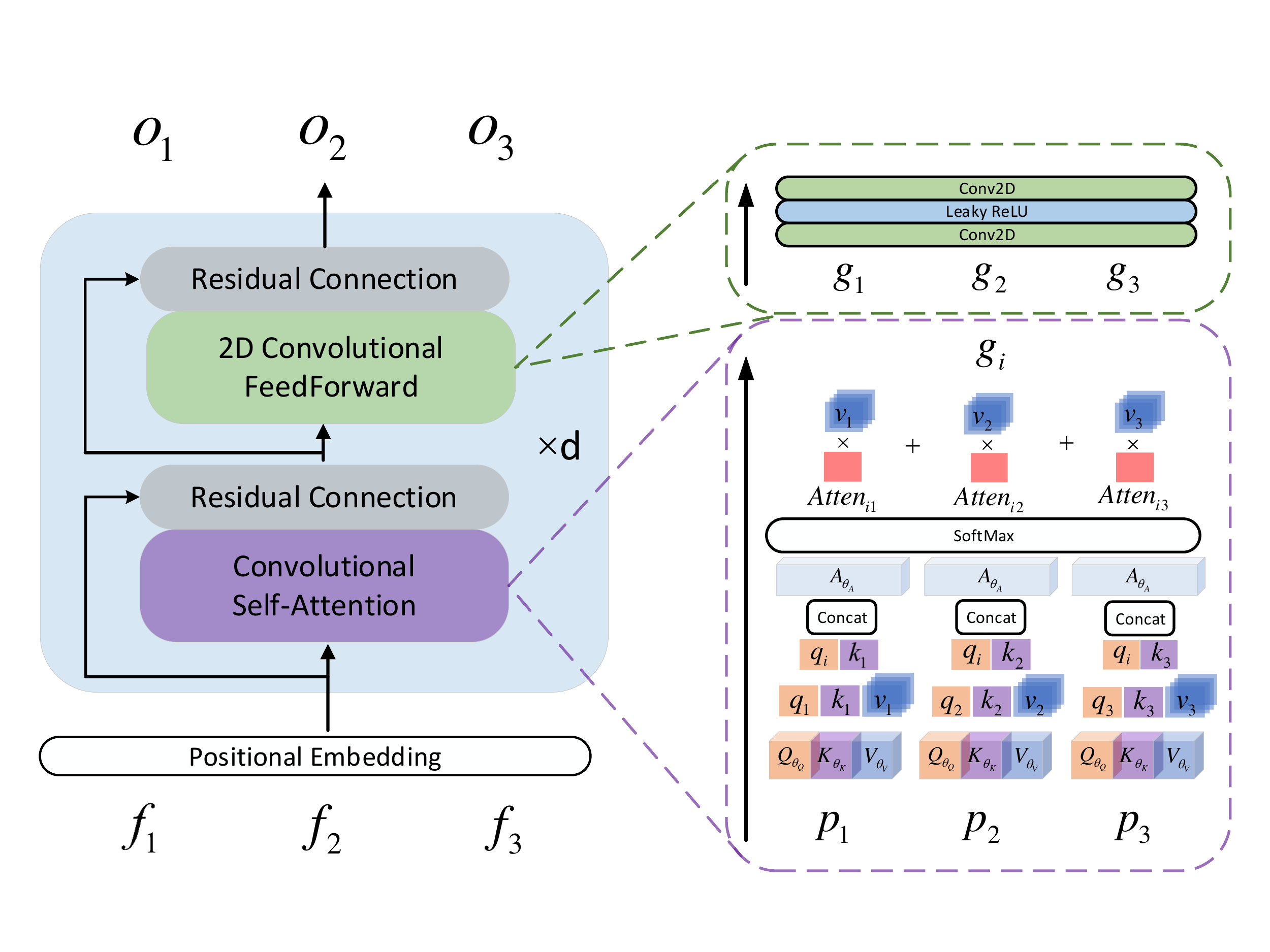}
    \caption{The Architecture of ConvTransformer.Queries, keys, and values are generated by 3 convolution sub-networks: $\mathcal{Q}_{\theta_{\mathcal{Q}}}$, $\mathcal{K}_{\theta_{\mathcal{K}}}$, and $\mathcal{V}_{\theta_{\mathcal{V}}}$. The attention of $q_i$ and $k_j$ is approximated by convolutional sub-network $\mathcal{A}_{\theta_{\mathcal{A}}}.$}
    \label{fig:ctrans}
\end{figure}

The sequence of feature maps with positional embeddings $p = \{p_0,p_1,\cdots,p_{N-1} \}$ where $p_i\in\mathbb{R}^{h\times w\times c} $ can be fed into the ConvTransformer. The ConvTransformer is a stack of several identical ConvTransformer layers. The architectures of ConvTransformer and its layers are shown in Fig.~\ref{fig:ctrans}. Each layer consists of 2 sub-layers: the convolutional self-attention layer and the feed-forward layer. When getting the sequence of input feature maps, three different convolution sub-networks $\mathcal{Q}_{\theta_{\mathcal{Q}}},$ $\mathcal{K}_{\theta_{\mathcal{K}}}$, and $\mathcal{V}_{\theta_{\mathcal{V}}}$ are used to generate the query sequence $q = \{q_0,q_1,\cdots,q_{N-1} \},q_i\in \mathbb{R}^{h\times w \times1}$, the key sequence $k = \{k_0,k_1,\cdots,k_{N-1} \},k_i\in \mathbb{R}^{h\times w \times1}$, and the value sequence $v = \{v_0,v_1,\cdots,v_{N-1}\},v_i\in \mathbb{R}^{h\times w \times c}$:
\begin{equation}
    \begin{gathered}
        q_i = \mathcal{Q}_{\theta_{\mathcal{Q}}}(p_i)\,,
        k_i = \mathcal{K}_{\theta_{\mathcal{K}}}(p_i)\,,
        v_i = \mathcal{V}_{\theta_{\mathcal{V}}}(p_i)\,, \\
        i\in[0,N-1]\,.
    \end{gathered}
\end{equation}
In this way, the attention map of two arbitrary frames $i$ and $j$ can be obtained by another convolution sub-network $\mathcal{A}_{\theta_{\mathcal{A}}}$:
\begin{gather}
    Atten(i,j) = \mathcal{A}_{\theta_{\mathcal{A}}}(concat[q_i,k_j])\,, \quad i,j\in[0,N-1]\,,
\end{gather}
where $concat$ refers to a concatenation operation of $q_i$ and $k_j$ in the channel dimension, $Atten(i,j)\in \mathbb{R}^{h\times w \times1}$.

For frame $i$, when getting attention maps of all input frames $j\in[0,N-1]$, a $SoftMax$ operation is applied to the attention maps. The output $g_i$ corresponding to feature map $f_i$ can be calculated as a weighted sum of all values:
\begin{gather}
    Atten(i,j) = SoftMax(Atten(i,j)),\quad j\in[0,N-1]\,, \\
    g_i = \sum_{j=0}^{N-1}Atten(i,j)v_j,\quad g_i\in \mathbb{R}^{h\times w\times c}\,.
\end{gather}

With the ConvTransformer architecture, the next step is to force the model to learn temporal correlations among input frames. We design our masked frames predicting strategy to build the inter-frame temporal correlations. The masked frames predicting strategy will be elaborated in the next section.

\subsection{Masked frames predicting}
\label{sec:fm}
With the ConvTransformer described in Sec.~\ref{sec:ctrans}, the next problem is how to enforce the model to build inter-frame temporal consistency. In this section, we will elaborate on our masked frames predicting strategy for learning video consistency.

Given a sequence of $N$ input video frames $F = \{F_0,F_1,\cdots,F_{N-1} \},$ we can get the corresponding sequence of feature maps with positional embeddings $p = \{p_0,p_1,\cdots,p_{N-1} \}$ as shown in Eq.~(\ref{equ:(1)}) and Eq.~(\ref{equ:(3)}). If we directly fed all the feature maps into ConvTransformer, the input would be highly redundant and the computational burden could be heavy. The high redundancy is not
beneficial for learning temporal consistency. Previous methods~\cite{mae,beit} reduce the redundancy of single images by masking a certain portion of patches. They force their model to predict the masked tokens or patches, and learn high-level semantic information rather than low-level details.
% taking $N=12$ as an example,

Inspired by the similarity between images spatial redundancy and videos temporal redundancy, we design our masked frames predicting strategy. For the spatial feature maps of $N$ input frames, we randomly mask a certain portion of them. In our case, considering the higher redundancy of videos, we adopt a very high masking ratio, which benefits the final results for both accuracy and consistency in our experiments. Retaining only a minority of the input frames ensures that we can represent inter-frame temporal correlations with lower redundancy. We illustrate our masking sampling strategies and masking ratios in Sec.~\ref{sec:im}. We also conduct ablation study on masking ratios in Sec.~\ref{sec:ab}.

The unmasked feature maps $p_{um}$ will be fed into the temporal structure encoder to build temporal correlations. The encoder is a six-layer ConvTransformer described in Sec.~\ref{sec:ctrans}. With the masking strategy, the video redundancy and computational cost are reduced. The encoder learns temporal correlations between the unmasked frames. If we denote the temporal structure encoder by $E_{\theta_{E}}$, the output temporal correlation features of unmasked frames $t_{um}$, where $t_{um}\subset \mathbb{R}^{h\times w\times c}$ can be expressed as:
\begin{equation}
    t_{um} = E_{\theta_E}(p_{um})\,.
\end{equation}

The next step is to complete the masked features according to the unmasked temporal correlation features $t_{um}$. We set a learnable mask token $m\in\mathbb{R}^{h\times w \times c}$ to represent the masked feature maps. The mask token is shared across all masked frames. We use the mask token to complete the corresponding positions of masked frames. The unmasked temporal correlation features $t_{um}$ remain on original positions. Positional embeddings as Eq.~(\ref{equ:pos}) and Eq.~(\ref{equ:(3)}) are added to the full sequence again because the original mask token does not have temporal information. We denote the full sequence with $m$, $t_{um}$, and positional embeddings by $t_{f}$. We use our temporal structure decoder $D_{\theta_D}$, a one-layer ConvTransformer, to reconstruct temporal features of masked positions based on the unmasked temporal correlation features:
\begin{equation}
    t_{r} = D_{\theta_D}(t_{f})\,,
\end{equation}
where $t_{r} = \{t_{r_1},t_{r_2},\cdots,t_{r_{N-1}}\}$ denotes the temporal structure features sequence of all input frames. In this way, the features of masked frames are completed according to the unmasked temporal correlation features. The video redundancy is reduced by masking and the temporal consistency is built by mask reconstruction. Experimental results show that our masked frames predicting strategy can efficiently improve the consistency of video depth results.

Finally, the last step is to recover depth maps from the temporal structure features sequence $t_{r}$. Our depth predictor is a CNN decoder with five up-projection modules to improve the spatial resolution and decrease the channel numbers. Skip connections and feature fusion model~\cite{FFM1,FFM2} are used to fuse the temporal structure features and the spatial structure features. Combining spatial and temporal information, the depth predictor can finally predict video depth results with both spatial accuracy and temporal consistency. With our masked frames predicting strategy illustrated, we will elaborate on the loss function in the next section.
\subsection{Training Loss}
\label{sec:de}
We adopt the widely-used scale-invariant loss~\cite{silog} to measure the discrepancy between output depth and ground truth. Given the predicted depth map for one frame $d\in\mathbb{R}^{H\times W}$ and the ground truth $d^*\in\mathbb{R}^{H\times W}$, the loss can be expressed as:
\begin{equation}
    \begin{gathered}
    L(u) = \alpha\sqrt{\frac{1}{n}\sum_iu_i^2 - \frac{\lambda}{n^2}(\sum_iu_i)^2}\,, \notag \\
    u_i = \log d_i - \log d_i^*\,,
    \end{gathered}
\end{equation}
where $n$ denotes the number of pixels with valid ground truth value and $i$ denotes pixels index. Similar to many other works, we use $\lambda = 0.85$ and $\alpha = 10$ in all our experiments.

Previous works utilize optical flow, camera poses, or GANs, which could fail when those extra information is inaccurate. Our method does not rely on additional information. Based on our masked frames predicting strategy, our \network{} can produce consistent video depth results only by the same training paradigm as single image depth estimation.

% \setlength{\abovecaptionskip}{+10pt}
% \begin{figure}
%     \centering
%     \includegraphics[scale=0.13,trim=110 0 110 10,clip]{figure/auau.png}
%     \caption{Motivation review. We divide a video clip into several $N$ frames sequences for inference. X-axis represents the frame numbers and Y-axis means our temporal consistency metric $OPW_t$ for each consecutive frame pair. At the junction of two subsequences, we can observe that the ST-CLSTM~\cite{ST-CLSTM} could cause obvious flickering and inconsistency due to its seriality and locality. Our \network{} has characteristic of parallelism and globality. We achieve better temporal consistency whether inside or between input sequences.}
%     \label{fig:jianci}
% \end{figure}
\section{EXPERIMENTS}
In this section, we evaluate our \network{} on the indoor NYU Depth V2~\cite{nyu} dataset and the outdoor KITTI~\cite{kitti} dataset.
% In section~\ref{},
Firstly,
we briefly introduce the datasets. In Sec.\ref{sec:mt}, we describe the evaluation metrics for spatial accuracy and temporal consistency. Some implementation details are shown in Sec.~\ref{sec:im}. We further illustrate our motivation by experiments %of video depth consistency
in Sec.~\ref{sec:mr}. The comparison results with state-of-the-art methods are shown in Sec.~\ref{sec:cp}. We conduct ablation studies to prove the effectiveness of our methods in Sec.~\ref{sec:ab}. We also compare the inference speed of different methods in Sec.~\ref{sec:speed}.
\subsection{Datasets}
\label{sec:data}
\textbf{NYU Depth V2}
contains 464 videos taken from indoor scenes. We apply the same train/test split as Eigen \textit{et al.}~\cite{silog} with 249 videos for training and 654 samples from the rest 215 videos for testing. We use the resolution of $640\times 480$ for training.

\textbf{KITTI} contains 61 outdoor video scenes captured by cameras
and depth sensors mounted on a driving car. We apply the same
train/test split as Eigen \textit{et al.}~\cite{silog} with 32 videos for
training and 697 samples from the rest 29 videos for testing. We use the resolution of $1216\times 352$ for training.
\subsection{Evaluation Metrics}
\label{sec:mt}
We evaluate the performance of our \network{} using depth estimation metrics %to measure the depth accuracy
and the optical flow consistency metric. %to measure the video depth temporal consistency.
We adopt the commonly applied depth metrics $Rel$, $RMSE$, $\log 10$, and $\delta_i (i=1,2,3)$.
% \begin{itemize}[leftmargin=*]
% \item Mean relative error (REL): $\frac{1}{n}\sum_{i=1}^n\frac{||d_i - d_i^*||_1}{d_i^*};$
% \item Root mean squared error (RMSE): $\sqrt{\frac{1}{n}\sum_{i=1}^n(d_i - d_i^*)^2};$
% \item Mean $\log_{10}$ error ($\log10$): $\frac{1}{n}\sum_{i=1}^n||\log_{10}d_i - \log_{10}d_i^*||_1;$
% \item Accuracy with threshold $t$: Percentage of $d_i$ such that $\\max(\frac{d_i}{d_i^*},\frac{d_i^*}{d_i}) = \delta<t\in\left[1.25, 1.25^2, 1.25^3\right],$
% \end{itemize}
% where $n$ denotes the total number of pixels, $d_i$ and $d_i^*$ are estimated and ground truth depth of pixel $i$, respectively.

\setlength{\abovecaptionskip}{+10pt}
\begin{figure}
    \centering
    \includegraphics[scale=0.13,trim=110 0 110 10,clip]{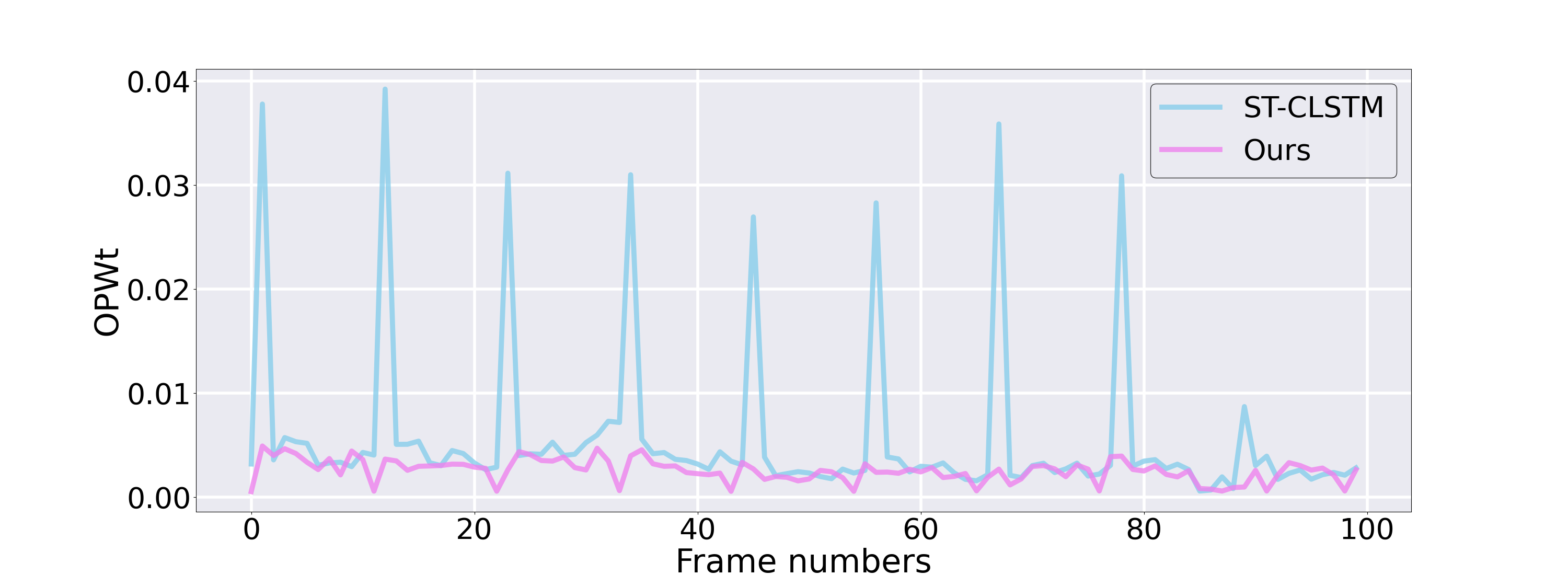}
    \caption{Motivation review. %We divide a video clip into several $N$ frames sequences for inference.
    The X-axis represents the frame numbers and Y-axis means our temporal consistency metric $\textbf{OPW}_{\textbf{t}}$ for each consecutive frame pair.}% At the junction of two sequences, we can observe that the ST-CLSTM~\cite{ST-CLSTM} causes obvious flickering and inconsistency due to its seriality and locality. Our \network{} has characteristics of parallelism and globality. We achieve better temporal consistency whether inside or between input sequences.}
    \label{fig:jianci}
\end{figure}

\begin{figure*}
    \centering
    \includegraphics[scale=0.130,trim=35 0 200 0,clip]{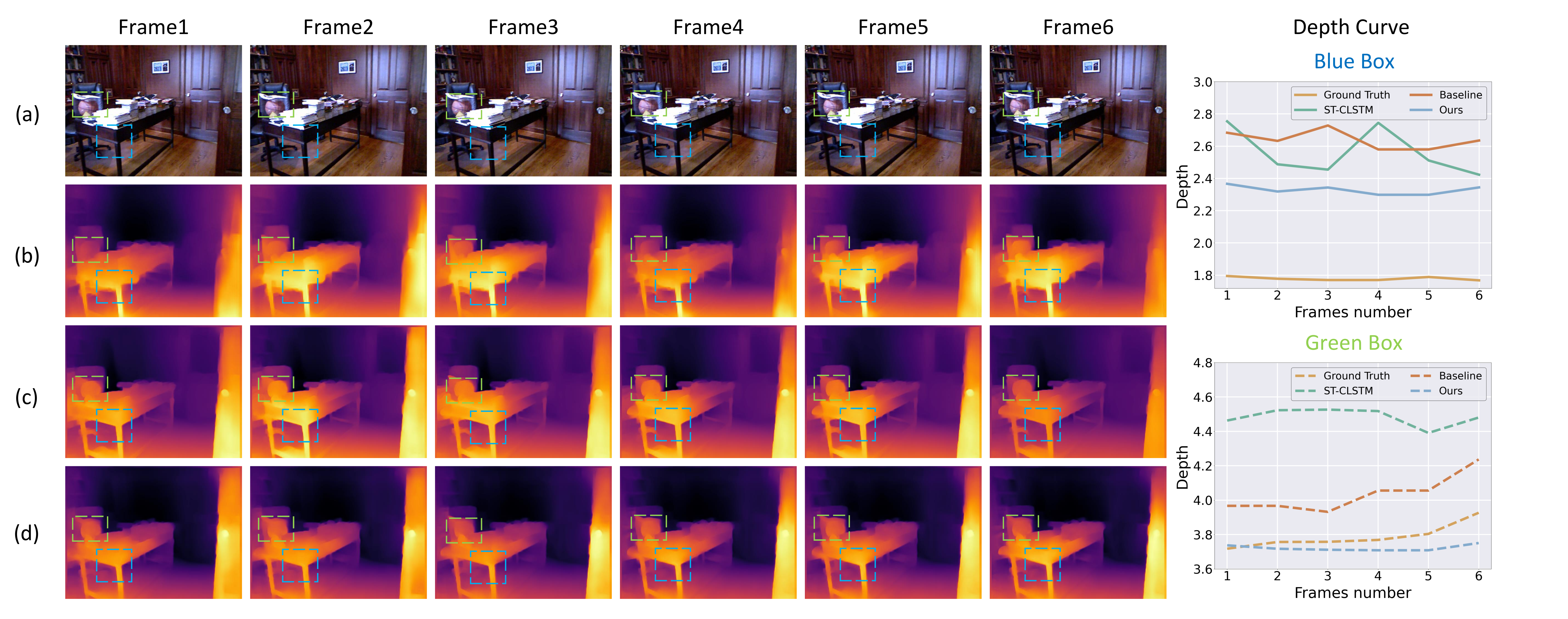}
\caption{Qualitative depth results on the NYU Depth V2 dataset. The four rows are: (a) RGB inputs; (b) Results of ST-CLSTM~\cite{ST-CLSTM}; (c) Baseline results; (d) Results of our FMNet. We highlight obviously different regions in dashed rectangular. For better comparison, we draw depth curves on the last column. Each curve represents depth value for the center point of a certain box in the six frames. Our FMNet shows higher depth accuracy and better temporal consistency than ST-CLSTM~\cite{ST-CLSTM} and our baseline.}
\label{fig:bigksh}
\end{figure*}

As for the temporal consistency metric, we adopt the optical flow based warping loss ($OPW$) proposed by~\cite{MM21}. They adopt it as a loss function for temporal consistency supervision. % in their approach.
In our case, we do not use it for training. Instead, we use $OPW$ to quantitatively evaluate the temporal consistency. % of video depth results.
The $OPW$ is defined as follows:
\begin{equation}
 \begin{gathered}
    OPW_{t} = \frac{1}{n}\sum_{i=1}^n M^{(i)}_{t+1 \Rightarrow t}||d^{(i)}_{t+1} - \hat{d}^{(i)}_{t}||_1\,, \\
    OPW = \sum_{t=0}^{T-1}OPW_{t}\,, \label{Eq:opl}
\end{gathered}
\end{equation}
where $\hat{d}_{t}$ is the predicted depth $d_{t}$ warped by the backward optical flow $FL_{t+1 \Rightarrow t}$ between input frames $F_t$ and $F_{t+1}.$ In our implementation, we adopt the trained RAFT~\cite{raft} model to compute optical flow between frames. For a video with $T$ frames, we calculate the $OPWt$ between each consecutive frame pair and add them together as the consistency index for the certain video. $M^{(i)}_{t+1 \Rightarrow t}$ is the visibility mask calculated from the warping discrepancy between frame $F_{t+1}$ and the warped frame $\hat{F}_t$:
\begin{equation}
    M^{(i)}_{t+1 \Rightarrow t} = \exp(-\beta||F_{t+1}-\hat{F}_t||_2^2)\,.
\end{equation}
Identical to~\cite{MM21}, we set $\beta = 50$ and use bilinear sampling layer~\cite{cz} for frames warping.

As for the evaluation on a test dataset with many testing videos, we calculate the $OPW$ as Eq.~(\ref{Eq:opl}) for each video and add them together as the final $OPW$ index on the test dataset.

\subsection{Implementation Details}
\label{sec:im}
We train our \network{} for 20 epochs on the NYU Depth V2 dataset. We use $0.0001$ as the initial learning rate, which decreases by a factor of 0.1 after every $5$ epochs. We cut videos to sequences with $N=12$ consecutive frames as input. As for the KITTI dataset, we train
our proposed \network{} for 30 epochs. The initial learning rate is also $0.0001$ and decreases by a factor of 0.1 after 10 epochs. We adopt the sequence length $N=8$. Without loss of generality, our spatial structure feature extractor is based on the ResNeXt-101~\cite{resnext}.%, which is widely used in many computer vision tasks.

As for the masking strategy in training time, we input $N$ frames each time and randomly mask $N - 2$ frames. The masking ratio is $83.33\%$ with $N=12$ on the NYU Depth V2 dataset and $75\%$ with $N=8$ on the KITTI dataset. The random masking strategy can be assumed as a form of data argumentation so that our model can learn inter-frame correlations with various time intervals. Therefore, we do not apply any other data augmentation methods. Considering the higher redundancy of videos than that of single images, we only retain $2$ frames for each input sequence. The redundancy will be minimized and the inter-frame temporal correlations still remain.

However, random masking strategy for inference could cause randomness in depth results. Instead, we use uniform masking for inference. For example, we will retain the fourth and eighth frames with $N=12$. We further ablate our masking ratios in Sec.~\ref{sec:ab}.

To prove the effectiveness of our approach, we also implement a baseline model without our temporal structure feature extractor. The baseline only consists of the spatial structure feature extractor and the depth predictor. It can be assumed as a CNN model for single image depth estimation without transformer and masking. %It is trained with the same loss function as our \network{}.
\subsection{Motivation Review}
\label{sec:mr}
In this section, we conduct an experiment to further expound on the motivation of our approach. In practice, a certain video clip will be partitioned into several video sequences with $N$ frames due to the limitation of computational resources. In our case, we set $N=12$ in Fig.~\ref{fig:jianci}. The traditional temporal models such as the ST-CLSTM~\cite{ST-CLSTM} process a certain input video sequence frame by frame. The inter-frame temporal correlations are built relying on the memory cell in a serial manner. We adopt both the ST-CLSTM~\cite{ST-CLSTM} and our ~\network{} to process the same input video clip which is divided into several $N$ frames sequences. We compare the temporal consistency metric $OPW_t$ for each consecutive frame pair.

At the junction of two sequences, we can observe that the ST-CLSTM~\cite{ST-CLSTM} causes obvious flickering and inconsistency due to the disabled memory cell with initialization value, which is a common phenomenon for all input video clips. Passing the memory cell value of the previous sequence to the next one is also unreasonable and could cause error accumulation. The temporal consistency is more attached to adjacent frames. Passing the memory cell value too far has no benefit for consistency because the video scene might change completely. In this way, ST-CLSTM~\cite{ST-CLSTM} inevitably causes the obvious inconsistency between adjacent sequences due to its seriality and locality.

In order to solve this problem, we design our \network{} based on ConvTransformer and masked frames predicting. The ConvTransformer can process input frames in a parallel manner. The masked frames predicting strategy is inspired by the high temporal redundancy of videos and the image patch masking strategy in recent MAE~\cite{mae}. Compared with previous methods, our model has better characteristics of parallelism and globality. The model is forced to predict the depth results based on all possibly relevant input frames. In this way, our \network{} is equipped with a larger
temporal receptive field and achieves better temporal consistency whether inside or between input sequences without relying on additional optical flow or camera poses.
\begin{table}
  \caption{Comparisons with state-of-the-art methods on the NYU Depth V2 dataset. $\delta_i$ means $\delta<1.25^i$. We show our results in the last row. Best performance is in boldface.}
  \label{tab:NYU depth}
  \resizebox{\columnwidth}{!}{
  \begin{tabular}{lcccccc}
    \toprule
    Method & Rel & RMSE & $\log 10$ & $\delta_1$ & $\delta_2$ & $\delta_3$ \\
    \midrule
    % DepthTransfer~\cite{n1} & $0.350$ & $1.200$ & $0.131$ & $-$ & $-$ & $-$ \\
    % Make3D~\cite{k1} & $0.349$ & $1.214$ & $-$ & $0.447$ & $0.745$ & $0.897$\\
    % Liu \textit{et al.}~\cite{n3} & $0.335$ & $1.060$ & $0.127$ & $-$ & $-$ & $-$\\
    % Li \textit{et al.}~\cite{n4} & $0.232$ & $0.821$ & $0.094$ & $0.621$ & $0.886$ & $0.968$\\
    % Liu \textit{et al.}~\cite{n5} & $0.230$ & $0.824$ & $0.095$ & $0.614$ & $0.883$ & $0.971$\\
    % Wang \textit{et al.}~\cite{n6} & $0.220$ & $0.824$ & $-$ & $0.605$ & $0.890$ & $0.970$\\
    % Liu \textit{et al.}~\cite{k3} & $0.213$ & $0.759$ & $0.087$ & $0.650$ & $0.906$ & $0.976$\\
    % Eigen \textit{et al.}~\cite{n8} & $0.158$ & $0.641$ & $-$ & $0.769$ & $0.950$ & $0.988$\\
    % Chakrabarti \textit{et al.}~\cite{n9} & $0.149$ & $0.620$ & $-$ & $0.806$ & $0.958$ & $0.987$\\
    % Li \textit{et al.}~\cite{n10} & $0.143$ & $0.635$ & $0.063$ & $0.788$ & $0.958$ & $0.991$\\
    % Ma \textit{et al.}~\cite{n11} & $0.143$ & $-$ & $-$ & $0.810$ & $0.959$ & $0.989$\\
    Laina \textit{et al.}~\cite{n12} & $0.127$ & $0.573$ & $0.055$ & $0.811$ & $0.953$ & $0.988$\\
    Pad-net~\cite{n13} & $0.120$ & $0.582$ & $0.055$ & $0.817$ & $0.954$ & $0.987$\\
    Cao \textit{et al.}~\cite{n14} & $0.141$ & $0.540$ & $0.060$ & $0.819$ & $0.965$ & $0.992$\\
    DORN~\cite{n15} & $\textbf{0.115}$ & $0.509$ & $\textbf{0.051}$ & $0.828$ & $0.965$ & $0.992$\\
    ST-CLSTM~\cite{ST-CLSTM} & $0.131$ & $0.571$ & $0.056$ & $0.833$ & $0.965$ & $0.991$\\
    Cao \textit{et al.}~\cite{MM21} & $0.131$ & $0.574$ & $0.056$ & $\textbf{0.835}$ & $0.965$ & $0.990$\\
    \midrule
    Ours & $0.134$ & $\textbf{0.452}$ & $0.056$ & $0.832$ & $\textbf{0.968}$ & $\textbf{0.992}$\\
    \bottomrule
\end{tabular}
}
\end{table}

\begin{table}
  \caption{Comparisons with state-of-the-art methods on the KITTI dataset. %The first five rows are the results of single image depth estimation methods. The following six rows are the results of video depth estimation methods.
  The last row shows our results. Best performance is in boldface.}
  \label{tab:kitti depth}
  \resizebox{\columnwidth}{!}{
  \begin{tabular}{lcccccc}
    \toprule
    Method & Rel & RMSE & $\log 10$ & $\delta_1$ & $\delta_2$ & $\delta_3$ \\
    \midrule
    % \midrule
    % Make3D~\cite{k1} & $0.280$ & $8.734$ & $-$ & $0.601$ & $0.820$ & $0.926$\\
    % Eigen \textit{et al.}~\cite{silog} & $0.190$ & $7.156$ & $-$ & $0.692$ & $0.899$ & $0.967$\\
    % Liu \textit{et al.}~\cite{k3} & $0.217$ & $6.986$ & $-$ & $0.647$ & $0.882$ & $0.961$\\
    % LRC~\cite{k4} & $0.114$ & $4.935$ & $-$ & $0.861$ & $0.949$ & $0.976$\\
    % Kuznietsov \textit{et al.}~\cite{k5} & $0.113$ & $4.621$ & $-$ & $0.862$ & $0.960$ & $0.986$\\
    % \midrule
    Mahjourian \textit{et al.}~\cite{k6} & $0.159$ & $5.912$ & $-$ & $0.784$ & $0.923$ & $0.970$\\
    Zhou \textit{et al.}~\cite{k7} & $0.143$ & $5.370$ & $-$ & $0.824$ & $0.937$ & $0.974$\\
    ST-CLSTM~\cite{ST-CLSTM} & $0.101$ & $4.137$ & $0.043$ & $\textbf{0.890}$ & $\textbf{0.970}$ & $0.989$\\
    Patil \textit{et al.}~\cite{k8} & $0.111$ & $4.650$ & $-$ & $0.883$ & $0.961$ & $0.982$\\
    CVD~\cite{cvd} & $0.130$ & $4.876$ & $-$ & $0.878$ & $0.946$ & $0.970$\\
    Cao \textit{et al.}~\cite{MM21} & $0.109$ & $4.366$ & $0.047$ & $0.872$ & $0.962$ & $0.986$\\
    \midrule
    Ours & $\textbf{0.099}$ & $\textbf{3.832}$ & $\textbf{0.042}$ & $0.886$ & $0.968$ & $\textbf{0.989}$\\
    \bottomrule
    \vspace{-20pt}
\end{tabular}
}
\end{table}

%\begin{table}
%  \caption{Comparisons of the temporal consistency metric OPL in NYU Depth V2 dataset. Best performance is in boldface.}
%  \label{tab:OPL}
  %\resizebox{0.95\columnwidth}{!}{
%  \begin{subtable}[t]{0.495\linewidth}
%  \caption{NYU Depth V2}\centering
%  \begin{tabular}{lc}
%    \toprule
%    Method & $OPL$ \\
%    \midrule
%    \midrule
%    Zhang \textit{et al.}~\cite{}& $12.159$ \\
%    Ours (baseline)& $6.828$ \\
%    Ours& $\textbf{6.425}$ \\
%    \bottomrule
%\end{tabular}
%\end{subtable}
%\begin{subtable}[t]{0.495\linewidth}\centering
%  \caption{Kitti}
%  \begin{tabular}{lc}
%   \toprule
%    Method & $OPL$ \\
%    \midrule
%    \midrule
%    Zhang \textit{et al.}~\cite{}& $-$ \\
%    Ours (baseline)& $44.179$ \\
%    Ours& $\textbf{30.596}$ \\
%    \bottomrule
%\end{tabular}
%\end{subtable}
%}
%\end{table}

\subsection{Comparisons with state-of-the-art results}
\label{sec:cp}
In this section, we evaluate our \network{} on the NYU Depth V2 dataset and the KITTI dataset. We compare our \network{} with some state-of-the-art results. The depth estimation results are reported in Table~\ref{tab:NYU depth} and Table~\ref{tab:kitti depth}. Our approach achieves comparable depth estimation accuracy on the NYU Depth V2 dataset and the KITTI dataset. For
some metrics, we outperform state-of-the-art methods based on generative adversarial networks, optical flow, or knowledge distillation. We only use the scale-invariant loss~\cite{silog} as supervision, which shows the effectiveness of our design.

To further prove the effectiveness of our \network{} in temporal consistency, we compare the consistency metric $OPW$ with previous state-of-the-art method~\cite{ST-CLSTM} in Table~\ref{tab:effd}. Our approach outperforms the ST-CLSTM~\cite{ST-CLSTM} by a large margin. The ST-CLSTM causes obvious flickering and inconsistency between adjacent sequences due to its seriality and locality. It can only produce depth results frame by frame. By contrast, our \network{} has better characteristic of parallelism and globality. It can produce depth results of all input frames in one time. In Fig.~\ref{fig:jianci}, we show $OPW_t$ values frame by frame. Our approach achieves better consistency whether inside or between input sequences. We also show some visual results of the NYU Depth V2 dataset in Fig.~\ref{fig:bigksh}. The visualization results and the depth curves show that our \network{} achieves video depth results with higher accuracy and better consistency.

CVD~\cite{cvd} is one of the most famous methods in consistent video depth estimation. Based on the trained Midas~\cite{midas} %or Monodepth2~\cite{monodepth2}
for single image depth estimation, CVD is in the test-time training paradigm. Their method highly relies on camera poses~\cite{colmapsfm,colmapmvs} and optical flow~\cite{flownet2}. When these information cannot be accurate in videos with dynamic scenes such as the KITTI dataset, CVD inevitably fails and causes large errors in depth results. They even need to finetune the optical flow model~\cite{flownet2} to get the depth metrics reported in Table~\ref{tab:kitti depth}. On the KITTI dataset, our method outperforms CVD in depth estimation metrics. We also achieve more than $30\%$ improvement of $OPW$ as shown in Table~\ref{tab:abma}. While CVD fails on the KITTI dataset, our method remains highly effective for consistent video depth estimation.

Some structure-from-motion (SFM) methods~\cite{banet,deepv2d} could achieve higher depth accuracy metrics. Those methods predict depth maps by feature matching over multiple frames. This idea benefits static scenes but does not account for dynamically moving objects. They need to mask the moving cars or people for pose estimation on the KITTI dataset. Those methods inevitably fail for videos with natural scenes or objects motion. By contrast, our method is not limited by camera poses. Our \network{} is also significantly faster than SFM-based methods due to the time-consuming pose estimation.
\begin{table}

  \caption{Effectiveness of our method in temporal consistency. We compare depth accuracy and temporal consistency of ST-CLSTM~\cite{ST-CLSTM} and our FMNet on the NYU depth V2 dataset. Our method outperforms the ST-CLSTM~\cite{ST-CLSTM} by a large margin in video depth consistency.} %of ST-CLSTM~\cite{ST-CLSTM}, our baseline, and our \network{}. The range of $\textbf{OPW}$
%   is different
  %differs between the two datasets
  %because of their different depth range.}
%   because of the different range of depth values on the two datasets.}
  \label{tab:effd}
  \resizebox{\columnwidth}{!}{
  \begin{tabular}{lccccccc}
    \toprule
    Method & Rel & RMSE & $\log 10$ & $\delta_1$ & $\delta_2$ & $\delta_3$ & $OPW$ \\
    \midrule
     ST-CLSTM~\cite{ST-CLSTM} & $\textbf{0.131}$ & $0.571$ & $0.056$ & $\textbf{0.833}$ & $0.965$ & $0.991$ & $12.159$\\
     Ours & $0.134$ & $\textbf{0.452}$ & $\textbf{0.056}$ & $0.832$ & $\textbf{0.968}$ & $\textbf{0.992}$ & $\textbf{6.425}$\\
    \bottomrule
\end{tabular}
}
\end{table}
\begin{table}
  \caption{Ablation study on the transformer and masking. We report the depth accuracy and temporal consistency on the KITTI dataset. We test the inference time on one GTX 1080Ti GPU with eight $640\times480$ frames as input.} %of ST-CLSTM~\cite{ST-CLSTM}, our baseline, and our \network{}. The range of $\textbf{OPW}$
%   is different
  %differs between the two datasets
  %because of their different depth range.}
%   because of the different range of depth values on the two datasets.}
  \label{tab:abma}
  \resizebox{\columnwidth}{!}{
  \begin{tabular}{lcccccccc}
    \toprule
    Method & RMSE & $\log 10$ & $\delta_1$ & $\delta_2$ & $\delta_3$ & $OPW$ &Time(s) \\
    \midrule
    % \multicolumn{8}{c}{NYU Depth V2} \\
    % \midrule
    %  ST-CLSTM~\cite{ST-CLSTM} & $\textbf{0.131}$ & $0.571$ & $0.056$ & $\textbf{0.833}$ & $0.965$ & $0.991$ & $12.159$\\
    %  Ours (baseline) & $0.138$ & $0.468$ & $0.059$ & $0.818$ & $0.965$ & $0.992$ & $6.828$\\
    %  Ours & $0.134$ & $\textbf{0.452}$ & $\textbf{0.056}$ & $0.832$ & $\textbf{0.968}$ & $\textbf{0.992}$ & $\textbf{6.425}$\\
    %\midrule
    %\multicolumn{8}{c}{KITTI} \\
    %\midrule
    baseline & $3.905$ & $0.044$ & $0.875$ & $0.965$ & $0.988$ & $44.179$ & $\textbf{3.22}$\\
    baseline+transformer & $3.877$ & $0.044$ & $0.881$ & $0.966$ & $0.989$ & $39.164$ & $4.77$\\
    baseline+transformer+masking & $\textbf{3.832}$ & $\textbf{0.042}$ & $\textbf{0.886}$ & $\textbf{0.968}$ & $\textbf{0.989}$& $\textbf{30.596}$ & $3.36$\\
    \bottomrule
\end{tabular}
}
\end{table}

%\begin{table}
  %\caption{Effectiveness of our method for temporal OFL metric. (a) contains OPL on NYU Depth V2 dataset; (b) contains OPL on Kitti dataset. Best performance is in boldface and we calculate the percentage of outperforming in the last row.}
  %\label{tab:effc}
  %\resizebox{0.95\columnwidth}{!}{
  %\begin{subtable}[t]{0.495\linewidth}
  %\caption{NYU Depth V2}\centering
  %\begin{tabular}{lc}
    %\toprule
    %Method & $OPL$ \\
    %\midrule
    %\midrule
    %Ours (baseline)& $6.828$ \\
    %Ours& $\textbf{6.425}$ \\
    %\midrule
    %percentage & $5.90\%$ \\
    %\bottomrule
%\end{tabular}
%\end{subtable}
%\begin{subtable}[t]{0.495\linewidth}\centering
%  \caption{Kitti}
%  \begin{tabular}{lc}
%    \toprule
%    Method & $OPL$ \\
%    \midrule
%    \midrule
%    Ours (baseline)& $44.179$ \\
%    Ours& $\textbf{30.596}$ \\
%    \midrule
%    percentage & $30.75\%$ \\
%    \bottomrule
%\end{tabular}
%\end{subtable}
%}
%\end{table}

\begin{figure}[t]
    \begin{minipage}[b]{0.95\linewidth}
      \centering
      \centerline{\includegraphics[width=1.05\linewidth,trim=115 10 55 100,clip]{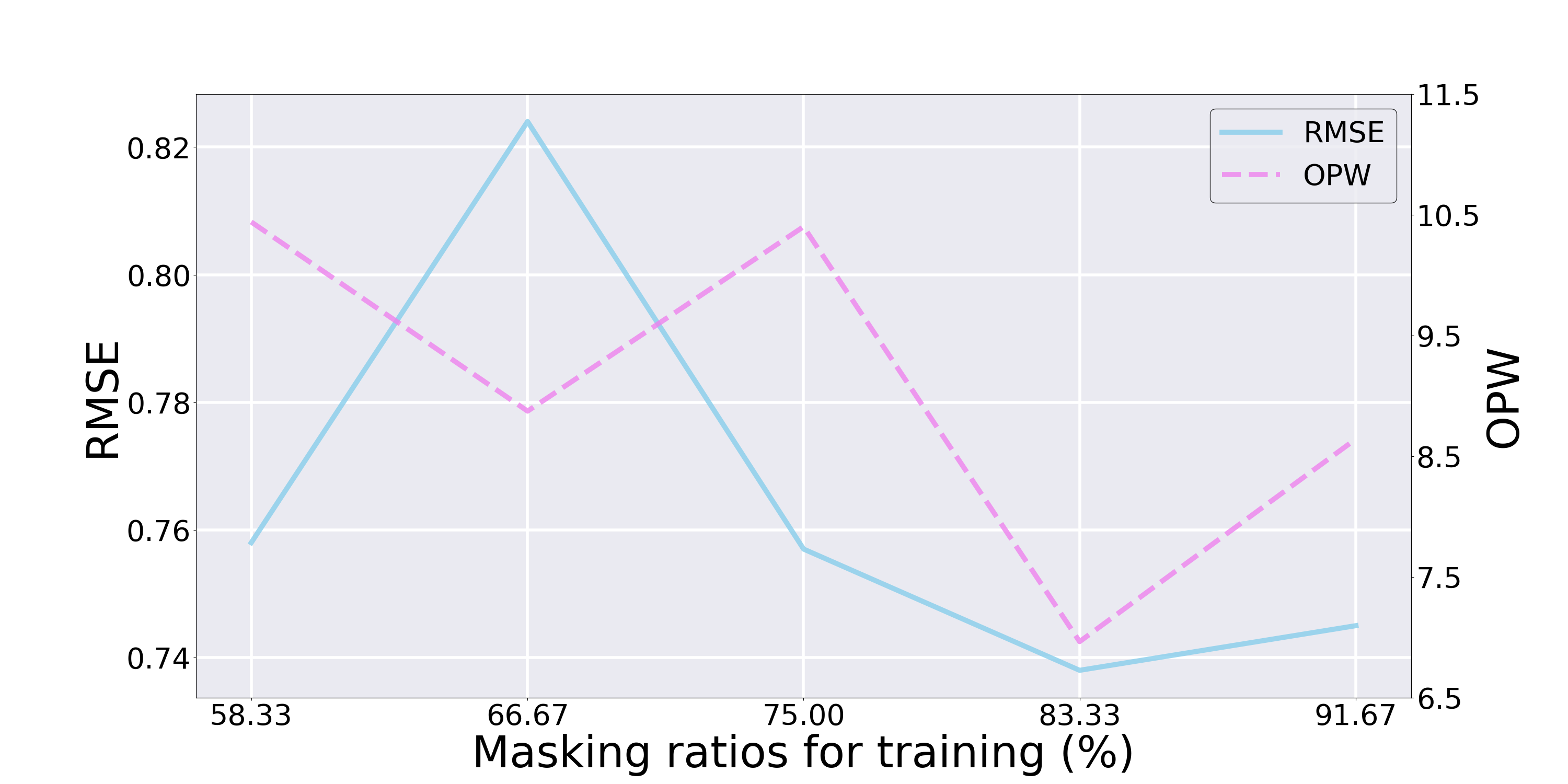}}
    \end{minipage}
\caption{Ablation study on masking ratios. The X-axis represents masking ratios and Y-axis means $\textbf{RMSE}$ and $\textbf{OPW}$. To reduce the experimental cost, we randomly choose $\textbf{40}$ videos for training and $\textbf{10}$ videos for $\textbf{OPW}$ evaluation on the NYU Depth V2 dataset. The %depth metric
$\textbf{RMSE}$ is evaluated on the public test dataset and cannot be compared with results in Table~\ref{tab:NYU depth}.}
\label{fig:ab}
\vspace{-10pt}
\end{figure}

% \begin{table}
%   \caption{Ablation study on our masking strategy and ratio. We use random mask for training and uniform mask for inference. N denotes the length of input video subsequences.The first two rows are the results for training and the last two rows are the results for inference. Best performance is in boldface.}
%   \label{tab:ab}

%   \resizebox{0.8\columnwidth}{!}{
%   \begin{tabular}{c|ccc}
%     \toprule
%     \multicolumn{4}{c}{Training} \\
%     \midrule
%     masked frames (random)& $N-2$ & $N-1$ & $N-3$\\
%     \midrule
%     $OPW$ & $\textbf{6.965}$ & $8.644$ & $10.402$\\
%     \midrule
%     \multicolumn{4}{c}{Inference} \\
%     \midrule
%     masked frames (uniform)& $N-2$ & $N-3$ & $0$\\
%     \midrule
%     $OPW$ & $\textbf{6.965}$ & $7.011$ & $7.192$\\
%     \bottomrule
% \end{tabular}
% }
% \end{table}

\subsection{Ablation studies}
\label{sec:ab}
% \subsubsection{Effectiveness of our method.}
% ST-CLSTM~\cite{ST-CLSTM} causes obvious flickering and inconsistency due to its seriality and locality. It can only produce depth results frame by frame. By contrast, our \network{} has better characteristic of parallelism and globality. It can produce depth results of all input frames in one time. In Table~\ref{tab:effd}, our FMNet achieves comparable depth accuracy and much better temporal consistency in comparison with ST-CLSTM~\cite{ST-CLSTM}.
\subsubsection{Transformer and masking.}
We further ablate the design of transformer and masking in Table~\ref{tab:abma}. The baseline is a CNN model without transformer and masking.  If we add the transformer to the baseline and directly input $N$ frames without masking, the depth accuracy and temporal consistency both improve but the inference speed decreases. With our masked frames predicting strategy, we force our model to learn the inter-frame temporal correlations.
Our \network{} achieves better temporal consistency, higher depth accuracy, and faster inference speed than the model without masking. The inference speed improves because the heavy computational burden of full sequences is only sustained by the lightweight one-layer temporal decoder.
%we force our model to learn the inter-frame temporal correlations. Compared with our baseline, our \network{} achieves %$5.90\%$ temporal consistency improvement on the NYU Depth V2 dataset
Compared with the baseline, our \network{} achieves $30.75\%$ temporal consistency improvement on the KITTI dataset. %As for the ST-CLSTM~\cite{ST-CLSTM}, our \network{} achieves comparable depth accuracy and much better temporal consistency whether inside or between input sequences.
\vspace{-3pt}
\subsubsection{Masking ratios for training}
The masking ratios significantly influence the temporal consistency. %We adopt the random masking for training, which can be assumed as a form of data argumentation. In this way, our model can learn temporal correlations with various time intervals.
We change the masking ratios for training. The experimental results are shown in Fig.~\ref{fig:ab}. The masking ratio of $83.33\%$ achieves the best accuracy and consistency. With 12 frames input and only 2 frames retained, the redundancy can be minimized while still preserving inter-frame correlations. If we use a higher masking ratio of $91.67\%$ with only one frame remaining, the redundancy is lower, however, the temporal correlations are lost. As for lower masking ratios, the redundancy is higher and the consistency is worse. We adopt the very high masking ratio with only 2 frames remaining in our approach.

\subsection{Inference speed comparison}
\label{sec:speed}
We test the inference time of DeepV2D~\cite{deepv2d}, CVD~\cite{cvd}, and Robust-CVD~\cite{rcvd} (officially published code). In Table~\ref{tab:speed}, we make a comparison with those methods on one GTX 1080Ti GPU with eight $640\times480$ frames as input. We can see that our FMNet, whether with ResNet18~\cite{resnet} or ResNext101~\cite{resnext} as the backbone, has significantly faster inference speeds than methods using pose estimation or optical flow such as DeepV2D~\cite{deepv2d}, CVD~\cite{cvd}, and Robust-CVD~\cite{rcvd}. %For DeepV2D with pre-computed camera poses, which means their motion module is not needed, DeepV2D is still significantly slower than our FMNet due to the time-consuming cost volumes and feature matching. Our \network{} is significantly faster than the test-time optimization methods CVD and Robust-CVD.
\begin{table}[h]
  \vspace{-15pt}
  \caption{Inference speed comparison. We test the speed using one GTX 1080Ti GPU with eight 640$\times$480 frames as input.}
  \label{tab:speed}
  \resizebox{\columnwidth}{!}{
  \begin{tabular}{lccc}
    \toprule
    Method & Camera pose & Optical flow & Time(s)\\
    \midrule
     Ours (ResNet18) & $\quad$ & $\quad$ & $1.32$\\
     Ours (ResNext101) & $\quad$ & $\quad$ & $3.36$\\
     DeepV2D (pose pre-computed)~\cite{deepv2d} & $\checkmark$ & $\quad$ & $11.65$\\
     DeepV2D (pose estimation)~\cite{deepv2d} & $\checkmark$ & $\quad$ & $55.62$\\
     CVD~\cite{cvd} & $\checkmark$ & $\checkmark$ & $376.32$\\
     Robust-CVD~\cite{rcvd} & $\checkmark$ & $\checkmark$ & $252.17$\\
    \bottomrule
    \vspace{-30pt}
\end{tabular}
}
\end{table}

% \begin{figure}
%     \centering
%     \includegraphics[scale=0.09,trim=60 0 0 0,clip]{figure/ab.pdf}
% \caption{Visual depth results comparison of ten and six masked frames for inference. The results are produced by our \network{} model on the whole NYU Depth V2 dataset, which is trained with $\textbf{N=12}$ input frames and $\textbf{N-2}$ random masking. With higher inference redundancy, the depth results of six masked frames have obviously worse consistency and flickering than ten masked frames.}
% \label{fig:abd}
% \end{figure}

% \subsubsection{Masking ratio for inference}
% We also ablate our masking ratio for inference in Fig.~\ref{fig:ab}(b). We use uniform mask for inference to avoid the randomness. For example with $N=12$ frames input, $N-2$ means that we retain the forth and eighth frames. For the same model trained with $N-2$ randomly masked frames, inferring with only less masked frames can cause the decrease of consistency due to higher redundancy. We also try the extreme situation: inference with no mask. We directly input $N$ frames to our temporal structure encoder. In this way, our model loses the vital mechanism of masked frames predicting which plays an vital role in temporal consistency. In Fig.~\ref{fig:abd}, based on $N=12$ and our \network{} trained on the whole NYU dataset in Table~\ref{tab:NYU depth}, we compare the depth results of $10$ and $6$ masked frames for inference. The visual depth results of $6$ masked frames have obviously worse consistency than $10$ masked frames due to higher redundancy.

\section{CONCLUSION}
% In this work, we provide a new perspective for consistent video depth estimation. Observing the redundancy of videos, we propose the frame masking network (FMNet) to mine temporal consistency based on the masking-and-reconstructing strategy. In our FMNet, input video frames are firstly encoded into spatial feature maps, a portion of which are then masked. The unmasked subset of the spatial feature maps are fed into the temporal structure encoder and are completed by a learnable mask token. The completed sequence is decoded into video depth results by our temporal structure decoder and depth predictor. In this manner, the FMNet learns the inter-frames temporal relation which derives more consistent video depth results compared with previous state-of-the-art approaches. Further analysis shows that with a very high masking ratio, the FMNet generates the most accurate and consistent results on the NYU Depth V2 dataset. The phenomenon reveals that the temporal consistency can be directly modeled from the videos rather than relying on additional information. Our work shows an undiscovered advantage of the masking-and-reconstructing strategy that it can help to generate consistent video depth results.\kexian{Please discuss the limitations in this section.}
In this work, we provide a new perspective for consistent video depth estimation. Observing the redundancy of videos,
we propose the frame masking network (FMNet) to mine consistency with the masking-and-reconstructing strategy. Randomly masking some input frames, we compel our model to estimate the depth of masked frames based on unmasked ones. %Our \network{} predicts depth results based on all possibly relevant input frames.
The implicit inter-frame correlations and the larger temporal receptive field lead to better temporal consistency compared with previous approaches. Further analysis shows that with a very high masking ratio, the FMNet generates the most accurate and consistent results on the NYU Depth V2 dataset. The phenomenon reveals that the consistency can be directly modeled from the videos. There are also some limitations of
our method. For example, we only use the scale-invariant loss as supervision. If ground truth exists flickering, it might lead to a measure of inconsistency in depth results. We can further introduce supervisions such as geometric constraints for this situation.

% \begin{acks}
% This work was funded by Adobe.
% \end{acks}
\section{ACKNOWLEDGMENTS}
This work was funded by Adobe.
%\newpage
\bibliographystyle{ACM-Reference-Format}
\bibliography{main}


\begin{thebibliography}{63}

%%% the Web address.

\ifx \showCODEN    \undefined \def \showCODEN     #1{\unskip}     \fi
\ifx \showDOI      \undefined \def \showDOI       #1{#1}\fi
\ifx \showISBNx    \undefined \def \showISBNx     #1{\unskip}     \fi
\ifx \showISBNxiii \undefined \def \showISBNxiii  #1{\unskip}     \fi
\ifx \showISSN     \undefined \def \showISSN      #1{\unskip}     \fi
\ifx \showLCCN     \undefined \def \showLCCN      #1{\unskip}     \fi
\ifx \shownote     \undefined \def \shownote      #1{#1}          \fi
\ifx \showarticletitle \undefined \def \showarticletitle #1{#1}   \fi
\ifx \showURL      \undefined \def \showURL       {\relax}        \fi
% invisible to TeX
\providecommand\bibfield[2]{#2}
\providecommand\bibinfo[2]{#2}
\providecommand\natexlab[1]{#1}
\providecommand\showeprint[2][]{arXiv:#2}

\bibitem[Arnab et~al\mbox{.}(2021)]%
        {vivit}
\bibfield{author}{\bibinfo{person}{Anurag Arnab}, \bibinfo{person}{Mostafa
  Dehghani}, \bibinfo{person}{Georg Heigold}, \bibinfo{person}{Chen Sun},
  \bibinfo{person}{Mario Lu{\v{c}}i{\'c}}, {and} \bibinfo{person}{Cordelia
  Schmid}.} \bibinfo{year}{2021}\natexlab{}.
\newblock \showarticletitle{Vivit: A video vision transformer}. In
  \bibinfo{booktitle}{\emph{Proceedings of the IEEE/CVF International
  Conference on Computer Vision (ICCV)}}. \bibinfo{pages}{6836--6846}.
\newblock


\bibitem[Bao et~al\mbox{.}(2022)]%
        {beit}
\bibfield{author}{\bibinfo{person}{Hangbo Bao}, \bibinfo{person}{Li Dong},
  \bibinfo{person}{Songhao Piao}, {and} \bibinfo{person}{Furu Wei}.}
  \bibinfo{year}{2022}\natexlab{}.
\newblock \showarticletitle{{BE}iT: {BERT} Pre-Training of Image Transformers}.
  In \bibinfo{booktitle}{\emph{International Conference on Learning
  Representations}}.
\newblock


\bibitem[Bhat et~al\mbox{.}(2021)]%
        {adabins}
\bibfield{author}{\bibinfo{person}{Shariq~Farooq Bhat},
  \bibinfo{person}{Ibraheem Alhashim}, {and} \bibinfo{person}{Peter Wonka}.}
  \bibinfo{year}{2021}\natexlab{}.
\newblock \showarticletitle{Adabins: Depth estimation using adaptive bins}. In
  \bibinfo{booktitle}{\emph{Proceedings of the IEEE/CVF Conference on Computer
  Vision and Pattern Recognition (CVPR)}}. \bibinfo{pages}{4009--4018}.
\newblock


\bibitem[Brown et~al\mbox{.}(2020)]%
        {gpt3}
\bibfield{author}{\bibinfo{person}{Tom Brown}, \bibinfo{person}{Benjamin Mann},
  \bibinfo{person}{Nick Ryder}, \bibinfo{person}{Melanie Subbiah},
  \bibinfo{person}{Jared~D Kaplan}, \bibinfo{person}{Prafulla Dhariwal},
  \bibinfo{person}{Arvind Neelakantan}, \bibinfo{person}{Pranav Shyam},
  \bibinfo{person}{Girish Sastry}, \bibinfo{person}{Amanda Askell},
  {et~al\mbox{.}}} \bibinfo{year}{2020}\natexlab{}.
\newblock \showarticletitle{Language models are few-shot learners}. In
  \bibinfo{booktitle}{\emph{Advances in neural information processing
  systems}}, Vol.~\bibinfo{volume}{33}. \bibinfo{pages}{1877--1901}.
\newblock


\bibitem[Cao et~al\mbox{.}(2021)]%
        {MM21}
\bibfield{author}{\bibinfo{person}{Yuanzhouhan Cao}, \bibinfo{person}{Yidong
  Li}, \bibinfo{person}{Haokui Zhang}, \bibinfo{person}{Chao Ren}, {and}
  \bibinfo{person}{Yifan Liu}.} \bibinfo{year}{2021}\natexlab{}.
\newblock \showarticletitle{Learning Structure Affinity for Video Depth
  Estimation}. In \bibinfo{booktitle}{\emph{Proceedings of the 29th ACM
  International Conference on Multimedia}}. \bibinfo{pages}{190--198}.
\newblock


\bibitem[Cao et~al\mbox{.}(2017)]%
        {n14}
\bibfield{author}{\bibinfo{person}{Yuanzhouhan Cao}, \bibinfo{person}{Zifeng
  Wu}, {and} \bibinfo{person}{Chunhua Shen}.} \bibinfo{year}{2017}\natexlab{}.
\newblock \showarticletitle{Estimating depth from monocular images as
  classification using deep fully convolutional residual networks}.
\newblock \bibinfo{journal}{\emph{IEEE Transactions on Circuits and Systems for
  Video Technology}} \bibinfo{volume}{28}, \bibinfo{number}{11}
  (\bibinfo{year}{2017}), \bibinfo{pages}{3174--3182}.
\newblock


\bibitem[Carion et~al\mbox{.}(2020)]%
        {detr}
\bibfield{author}{\bibinfo{person}{Nicolas Carion}, \bibinfo{person}{Francisco
  Massa}, \bibinfo{person}{Gabriel Synnaeve}, \bibinfo{person}{Nicolas
  Usunier}, \bibinfo{person}{Alexander Kirillov}, {and} \bibinfo{person}{Sergey
  Zagoruyko}.} \bibinfo{year}{2020}\natexlab{}.
\newblock \showarticletitle{End-to-end object detection with transformers}. In
  \bibinfo{booktitle}{\emph{European Conference on Computer Vision (ECCV)}},
  Vol.~\bibinfo{volume}{12346}. Springer, \bibinfo{pages}{213--229}.
\newblock


\bibitem[Chen et~al\mbox{.}(2020)]%
        {igpt}
\bibfield{author}{\bibinfo{person}{Mark Chen}, \bibinfo{person}{Alec Radford},
  \bibinfo{person}{Rewon Child}, \bibinfo{person}{Jeffrey Wu},
  \bibinfo{person}{Heewoo Jun}, \bibinfo{person}{David Luan}, {and}
  \bibinfo{person}{Ilya Sutskever}.} \bibinfo{year}{2020}\natexlab{}.
\newblock \showarticletitle{Generative pretraining from pixels}. In
  \bibinfo{booktitle}{\emph{International conference on machine learning}}.
  PMLR, \bibinfo{pages}{1691--1703}.
\newblock


\bibitem[Cheng et~al\mbox{.}(2018)]%
        {spn}
\bibfield{author}{\bibinfo{person}{Xinjing Cheng}, \bibinfo{person}{Peng Wang},
  {and} \bibinfo{person}{Ruigang Yang}.} \bibinfo{year}{2018}\natexlab{}.
\newblock \showarticletitle{Depth estimation via affinity learned with
  convolutional spatial propagation network}. In
  \bibinfo{booktitle}{\emph{European Conference on Computer Vision (ECCV)}},
  Vol.~\bibinfo{volume}{11220}. \bibinfo{pages}{108--125}.
\newblock


\bibitem[Devlin et~al\mbox{.}(2019)]%
        {bert}
\bibfield{author}{\bibinfo{person}{Jacob Devlin}, \bibinfo{person}{Ming-Wei
  Chang}, \bibinfo{person}{Kenton Lee}, {and} \bibinfo{person}{Kristina
  Toutanova}.} \bibinfo{year}{2019}\natexlab{}.
\newblock \showarticletitle{Bert: Pre-training of deep bidirectional
  transformers for language understanding}. In
  \bibinfo{booktitle}{\emph{Proceedings of the 2019 Conference of the North
  American Chapter of the Association for Computational Linguistics: Human
  Language Technologies,{NAACL-HLT} 2019, Minneapolis, MN, USA, June 2-7, 2019,
  Volume 1 (Long and Short Papers)}}. \bibinfo{pages}{4171--4186}.
\newblock


\bibitem[Dosovitskiy et~al\mbox{.}(2020)]%
        {VIT}
\bibfield{author}{\bibinfo{person}{Alexey Dosovitskiy}, \bibinfo{person}{Lucas
  Beyer}, \bibinfo{person}{Alexander Kolesnikov}, \bibinfo{person}{Dirk
  Weissenborn}, \bibinfo{person}{Xiaohua Zhai}, \bibinfo{person}{Thomas
  Unterthiner}, \bibinfo{person}{Mostafa Dehghani}, \bibinfo{person}{Matthias
  Minderer}, \bibinfo{person}{Georg Heigold}, \bibinfo{person}{Sylvain Gelly},
  {et~al\mbox{.}}} \bibinfo{year}{2020}\natexlab{}.
\newblock \showarticletitle{An Image is Worth 16x16 Words: Transformers for
  Image Recognition at Scale}. In \bibinfo{booktitle}{\emph{International
  Conference on Learning Representations}}.
\newblock


\bibitem[Eigen et~al\mbox{.}(2014)]%
        {silog}
\bibfield{author}{\bibinfo{person}{David Eigen}, \bibinfo{person}{Christian
  Puhrsch}, {and} \bibinfo{person}{Rob Fergus}.}
  \bibinfo{year}{2014}\natexlab{}.
\newblock \showarticletitle{Depth map prediction from a single image using a
  multi-scale deep network}. In \bibinfo{booktitle}{\emph{Advances in neural
  information processing systems}}, Vol.~\bibinfo{volume}{27}.
  \bibinfo{pages}{2366--2374}.
\newblock


\bibitem[Fan et~al\mbox{.}(2021)]%
        {mvit}
\bibfield{author}{\bibinfo{person}{Haoqi Fan}, \bibinfo{person}{Bo Xiong},
  \bibinfo{person}{Karttikeya Mangalam}, \bibinfo{person}{Yanghao Li},
  \bibinfo{person}{Zhicheng Yan}, \bibinfo{person}{Jitendra Malik}, {and}
  \bibinfo{person}{Christoph Feichtenhofer}.} \bibinfo{year}{2021}\natexlab{}.
\newblock \showarticletitle{Multiscale vision transformers}. In
  \bibinfo{booktitle}{\emph{Proceedings of the IEEE/CVF International
  Conference on Computer Vision (ICCV)}}. \bibinfo{pages}{6824--6835}.
\newblock


\bibitem[Fu et~al\mbox{.}(2018)]%
        {n15}
\bibfield{author}{\bibinfo{person}{Huan Fu}, \bibinfo{person}{Mingming Gong},
  \bibinfo{person}{Chaohui Wang}, \bibinfo{person}{Kayhan Batmanghelich}, {and}
  \bibinfo{person}{Dacheng Tao}.} \bibinfo{year}{2018}\natexlab{}.
\newblock \showarticletitle{Deep ordinal regression network for monocular depth
  estimation}. In \bibinfo{booktitle}{\emph{Proceedings of the IEEE/CVF
  Conference on Computer Vision and Pattern Recognition (CVPR)}}.
  \bibinfo{pages}{2002--2011}.
\newblock


\bibitem[Geiger et~al\mbox{.}(2013)]%
        {kitti}
\bibfield{author}{\bibinfo{person}{Andreas Geiger}, \bibinfo{person}{Philip
  Lenz}, \bibinfo{person}{Christoph Stiller}, {and} \bibinfo{person}{Raquel
  Urtasun}.} \bibinfo{year}{2013}\natexlab{}.
\newblock \showarticletitle{Vision meets robotics: The kitti dataset}.
\newblock \bibinfo{journal}{\emph{The International Journal of Robotics
  Research}} \bibinfo{volume}{32}, \bibinfo{number}{11} (\bibinfo{year}{2013}),
  \bibinfo{pages}{1231--1237}.
\newblock


\bibitem[Godard et~al\mbox{.}(2019)]%
        {monodepth2}
\bibfield{author}{\bibinfo{person}{C. Godard}, \bibinfo{person}{O. Aodha},
  \bibinfo{person}{M. Firman}, {and} \bibinfo{person}{G. Brostow}.}
  \bibinfo{year}{2019}\natexlab{}.
\newblock \showarticletitle{Digging into self-supervised monocular depth
  estimation}. In \bibinfo{booktitle}{\emph{Proceedings of the IEEE/CVF
  International Conference on Computer Vision (ICCV)}}.
  \bibinfo{pages}{3828--3838}.
\newblock


\bibitem[Goodfellow et~al\mbox{.}(2014)]%
        {gan}
\bibfield{author}{\bibinfo{person}{Ian Goodfellow}, \bibinfo{person}{Jean
  Pouget-Abadie}, \bibinfo{person}{Mehdi Mirza}, \bibinfo{person}{Bing Xu},
  \bibinfo{person}{David Warde-Farley}, \bibinfo{person}{Sherjil Ozair},
  \bibinfo{person}{Aaron Courville}, {and} \bibinfo{person}{Yoshua Bengio}.}
  \bibinfo{year}{2014}\natexlab{}.
\newblock \showarticletitle{Generative adversarial nets}. In
  \bibinfo{booktitle}{\emph{Advances in neural information processing
  systems}}, Vol.~\bibinfo{volume}{27}.
\newblock


\bibitem[He et~al\mbox{.}(2021)]%
        {mae}
\bibfield{author}{\bibinfo{person}{Kaiming He}, \bibinfo{person}{Xinlei Chen},
  \bibinfo{person}{Saining Xie}, \bibinfo{person}{Yanghao Li},
  \bibinfo{person}{Piotr Doll{\'a}r}, {and} \bibinfo{person}{Ross Girshick}.}
  \bibinfo{year}{2021}\natexlab{}.
\newblock \showarticletitle{Masked autoencoders are scalable vision learners}.
\newblock \bibinfo{journal}{\emph{arXiv preprint arXiv:2111.06377}}
  (\bibinfo{year}{2021}).
\newblock


\bibitem[He et~al\mbox{.}(2016)]%
        {resnet}
\bibfield{author}{\bibinfo{person}{Kaiming He}, \bibinfo{person}{Xiangyu
  Zhang}, \bibinfo{person}{Shaoqing Ren}, {and} \bibinfo{person}{Jian Sun}.}
  \bibinfo{year}{2016}\natexlab{}.
\newblock \showarticletitle{Deep residual learning for image recognition}. In
  \bibinfo{booktitle}{\emph{Proceedings of the IEEE/CVF Conference on Computer
  Vision and Pattern Recognition (CVPR)}}. \bibinfo{pages}{770--778}.
\newblock


\bibitem[Hinton et~al\mbox{.}(2015)]%
        {kd0}
\bibfield{author}{\bibinfo{person}{Geoffrey Hinton}, \bibinfo{person}{Oriol
  Vinyals}, \bibinfo{person}{Jeff Dean}, {et~al\mbox{.}}}
  \bibinfo{year}{2015}\natexlab{}.
\newblock \showarticletitle{Distilling the knowledge in a neural network}.
\newblock \bibinfo{journal}{\emph{arXiv preprint arXiv:1503.02531}}
  \bibinfo{volume}{2}, \bibinfo{number}{7} (\bibinfo{year}{2015}).
\newblock


\bibitem[Hochreiter and Schmidhuber(1997)]%
        {lstm}
\bibfield{author}{\bibinfo{person}{Sepp Hochreiter} {and}
  \bibinfo{person}{J{\"u}rgen Schmidhuber}.} \bibinfo{year}{1997}\natexlab{}.
\newblock \showarticletitle{Long short-term memory}.
\newblock \bibinfo{journal}{\emph{Neural computation}} \bibinfo{volume}{9},
  \bibinfo{number}{8} (\bibinfo{year}{1997}), \bibinfo{pages}{1735--1780}.
\newblock


\bibitem[Ilg et~al\mbox{.}(2017)]%
        {flownet2}
\bibfield{author}{\bibinfo{person}{Eddy Ilg}, \bibinfo{person}{Nikolaus Mayer},
  \bibinfo{person}{Tonmoy Saikia}, \bibinfo{person}{Margret Keuper},
  \bibinfo{person}{Alexey Dosovitskiy}, {and} \bibinfo{person}{Thomas Brox}.}
  \bibinfo{year}{2017}\natexlab{}.
\newblock \showarticletitle{Flownet 2.0: Evolution of optical flow estimation
  with deep networks}. In \bibinfo{booktitle}{\emph{Proceedings of the IEEE/CVF
  Conference on Computer Vision and Pattern Recognition (CVPR)}}.
  \bibinfo{pages}{2462--2470}.
\newblock


\bibitem[Jaderberg et~al\mbox{.}(2015)]%
        {cz}
\bibfield{author}{\bibinfo{person}{Max Jaderberg}, \bibinfo{person}{Karen
  Simonyan}, \bibinfo{person}{Andrew Zisserman}, {et~al\mbox{.}}}
  \bibinfo{year}{2015}\natexlab{}.
\newblock \showarticletitle{Spatial transformer networks}. In
  \bibinfo{booktitle}{\emph{Advances in neural information processing
  systems}}, Vol.~\bibinfo{volume}{28}.
\newblock


\bibitem[Karsch et~al\mbox{.}(2014)]%
        {n1}
\bibfield{author}{\bibinfo{person}{Kevin Karsch}, \bibinfo{person}{Ce Liu},
  {and} \bibinfo{person}{Sing~Bing Kang}.} \bibinfo{year}{2014}\natexlab{}.
\newblock \showarticletitle{Depth transfer: Depth extraction from video using
  non-parametric sampling}.
\newblock \bibinfo{journal}{\emph{IEEE transactions on pattern analysis and
  machine intelligence}} \bibinfo{volume}{36}, \bibinfo{number}{11}
  (\bibinfo{year}{2014}), \bibinfo{pages}{2144--2158}.
\newblock


\bibitem[Kopf et~al\mbox{.}(2021)]%
        {rcvd}
\bibfield{author}{\bibinfo{person}{Johannes Kopf}, \bibinfo{person}{Xuejian
  Rong}, {and} \bibinfo{person}{Jia-Bin Huang}.}
  \bibinfo{year}{2021}\natexlab{}.
\newblock \showarticletitle{Robust consistent video depth estimation}. In
  \bibinfo{booktitle}{\emph{Proceedings of the IEEE/CVF Conference on Computer
  Vision and Pattern Recognition (CVPR)}}. \bibinfo{pages}{1611--1621}.
\newblock


\bibitem[Laina et~al\mbox{.}(2016)]%
        {n12}
\bibfield{author}{\bibinfo{person}{Iro Laina}, \bibinfo{person}{Christian
  Rupprecht}, \bibinfo{person}{Vasileios Belagiannis},
  \bibinfo{person}{Federico Tombari}, {and} \bibinfo{person}{Nassir Navab}.}
  \bibinfo{year}{2016}\natexlab{}.
\newblock \showarticletitle{Deeper depth prediction with fully convolutional
  residual networks}. In \bibinfo{booktitle}{\emph{2016 Fourth international
  conference on 3D vision (3DV)}}. IEEE, \bibinfo{pages}{239--248}.
\newblock


\bibitem[LeCun et~al\mbox{.}(1998)]%
        {lenet}
\bibfield{author}{\bibinfo{person}{Yann LeCun}, \bibinfo{person}{L{\'e}on
  Bottou}, \bibinfo{person}{Yoshua Bengio}, {and} \bibinfo{person}{Patrick
  Haffner}.} \bibinfo{year}{1998}\natexlab{}.
\newblock \showarticletitle{Gradient-based learning applied to document
  recognition}.
\newblock \bibinfo{journal}{\emph{Proc. IEEE}} \bibinfo{volume}{86},
  \bibinfo{number}{11} (\bibinfo{year}{1998}), \bibinfo{pages}{2278--2324}.
\newblock


\bibitem[Lee et~al\mbox{.}(2019)]%
        {bts}
\bibfield{author}{\bibinfo{person}{Jin~Han Lee}, \bibinfo{person}{Myung-Kyu
  Han}, \bibinfo{person}{Dong~Wook Ko}, {and} \bibinfo{person}{Il~Hong Suh}.}
  \bibinfo{year}{2019}\natexlab{}.
\newblock \showarticletitle{From big to small: Multi-scale local planar
  guidance for monocular depth estimation}.
\newblock \bibinfo{journal}{\emph{arXiv preprint arXiv:1907.10326}}
  (\bibinfo{year}{2019}).
\newblock


\bibitem[Li et~al\mbox{.}(2018)]%
        {DABC}
\bibfield{author}{\bibinfo{person}{Ruibo Li}, \bibinfo{person}{Ke Xian},
  \bibinfo{person}{Chunhua Shen}, \bibinfo{person}{Zhiguo Cao},
  \bibinfo{person}{Hao Lu}, {and} \bibinfo{person}{Lingxiao Hang}.}
  \bibinfo{year}{2018}\natexlab{}.
\newblock \showarticletitle{Deep attention-based classification network for
  robust depth prediction}. In \bibinfo{booktitle}{\emph{Asian Conference on
  Computer Vision (ACCV)}}. \bibinfo{pages}{663--678}.
\newblock


\bibitem[Lin et~al\mbox{.}(2017b)]%
        {FFM1}
\bibfield{author}{\bibinfo{person}{Guosheng Lin}, \bibinfo{person}{Anton
  Milan}, \bibinfo{person}{Chunhua Shen}, {and} \bibinfo{person}{Ian Reid}.}
  \bibinfo{year}{2017}\natexlab{b}.
\newblock \showarticletitle{Refinenet: Multi-path refinement networks for
  high-resolution semantic segmentation}. In
  \bibinfo{booktitle}{\emph{Proceedings of the IEEE/CVF Conference on Computer
  Vision and Pattern Recognition (CVPR)}}. \bibinfo{pages}{1925--1934}.
\newblock


\bibitem[Lin et~al\mbox{.}(2017a)]%
        {FFM2}
\bibfield{author}{\bibinfo{person}{Tsung-Yi Lin}, \bibinfo{person}{Piotr
  Doll{\'a}r}, \bibinfo{person}{Ross Girshick}, \bibinfo{person}{Kaiming He},
  \bibinfo{person}{Bharath Hariharan}, {and} \bibinfo{person}{Serge Belongie}.}
  \bibinfo{year}{2017}\natexlab{a}.
\newblock \showarticletitle{Feature pyramid networks for object detection}. In
  \bibinfo{booktitle}{\emph{Proceedings of the IEEE/CVF Conference on Computer
  Vision and Pattern Recognition (CVPR)}}. \bibinfo{pages}{2117--2125}.
\newblock


\bibitem[Liu et~al\mbox{.}(2019)]%
        {kd1}
\bibfield{author}{\bibinfo{person}{Yifan Liu}, \bibinfo{person}{Ke Chen},
  \bibinfo{person}{Chris Liu}, \bibinfo{person}{Zengchang Qin},
  \bibinfo{person}{Zhenbo Luo}, {and} \bibinfo{person}{Jingdong Wang}.}
  \bibinfo{year}{2019}\natexlab{}.
\newblock \showarticletitle{Structured knowledge distillation for semantic
  segmentation}. In \bibinfo{booktitle}{\emph{Proceedings of the IEEE/CVF
  Conference on Computer Vision and Pattern Recognition (CVPR)}}.
  \bibinfo{pages}{2604--2613}.
\newblock


\bibitem[Liu et~al\mbox{.}(2020b)]%
        {kd2}
\bibfield{author}{\bibinfo{person}{Yifan Liu}, \bibinfo{person}{Changyong Shu},
  \bibinfo{person}{Jingdong Wang}, {and} \bibinfo{person}{Chunhua Shen}.}
  \bibinfo{year}{2020}\natexlab{b}.
\newblock \showarticletitle{Structured knowledge distillation for dense
  prediction}.
\newblock \bibinfo{journal}{\emph{IEEE transactions on pattern analysis and
  machine intelligence}} (\bibinfo{year}{2020}).
\newblock


\bibitem[Liu et~al\mbox{.}(2021)]%
        {swin}
\bibfield{author}{\bibinfo{person}{Ze Liu}, \bibinfo{person}{Yutong Lin},
  \bibinfo{person}{Yue Cao}, \bibinfo{person}{Han Hu}, \bibinfo{person}{Yixuan
  Wei}, \bibinfo{person}{Zheng Zhang}, \bibinfo{person}{Stephen Lin}, {and}
  \bibinfo{person}{Baining Guo}.} \bibinfo{year}{2021}\natexlab{}.
\newblock \showarticletitle{Swin transformer: Hierarchical vision transformer
  using shifted windows}. In \bibinfo{booktitle}{\emph{Proceedings of the
  IEEE/CVF International Conference on Computer Vision (ICCV)}}.
  \bibinfo{pages}{10012--10022}.
\newblock


\bibitem[Liu et~al\mbox{.}(2020a)]%
        {ctrans}
\bibfield{author}{\bibinfo{person}{Zhouyong Liu}, \bibinfo{person}{Shun Luo},
  \bibinfo{person}{Wubin Li}, \bibinfo{person}{Jingben Lu},
  \bibinfo{person}{Yufan Wu}, \bibinfo{person}{Shilei Sun},
  \bibinfo{person}{Chunguo Li}, {and} \bibinfo{person}{Luxi Yang}.}
  \bibinfo{year}{2020}\natexlab{a}.
\newblock \showarticletitle{Convtransformer: A convolutional transformer
  network for video frame synthesis}.
\newblock \bibinfo{journal}{\emph{arXiv preprint arXiv:2011.10185}}
  (\bibinfo{year}{2020}).
\newblock


\bibitem[Luo et~al\mbox{.}(2020)]%
        {cvd}
\bibfield{author}{\bibinfo{person}{Xuan Luo}, \bibinfo{person}{Jia-Bin Huang},
  \bibinfo{person}{Richard Szeliski}, \bibinfo{person}{Kevin Matzen}, {and}
  \bibinfo{person}{Johannes Kopf}.} \bibinfo{year}{2020}\natexlab{}.
\newblock \showarticletitle{Consistent video depth estimation}.
\newblock \bibinfo{journal}{\emph{ACM Transactions on Graphics (ToG)}}
  \bibinfo{volume}{39}, \bibinfo{number}{4} (\bibinfo{year}{2020}),
  \bibinfo{pages}{71--1}.
\newblock


\bibitem[Mahjourian et~al\mbox{.}(2018)]%
        {k6}
\bibfield{author}{\bibinfo{person}{Reza Mahjourian}, \bibinfo{person}{Martin
  Wicke}, {and} \bibinfo{person}{Anelia Angelova}.}
  \bibinfo{year}{2018}\natexlab{}.
\newblock \showarticletitle{Unsupervised learning of depth and ego-motion from
  monocular video using 3d geometric constraints}. In
  \bibinfo{booktitle}{\emph{Proceedings of the IEEE/CVF Conference on Computer
  Vision and Pattern Recognition (CVPR)}}. \bibinfo{pages}{5667--5675}.
\newblock


\bibitem[Patil et~al\mbox{.}(2020)]%
        {k8}
\bibfield{author}{\bibinfo{person}{Vaishakh Patil}, \bibinfo{person}{Wouter
  Van~Gansbeke}, \bibinfo{person}{Dengxin Dai}, {and} \bibinfo{person}{Luc
  Van~Gool}.} \bibinfo{year}{2020}\natexlab{}.
\newblock \showarticletitle{Don’t forget the past: Recurrent depth estimation
  from monocular video}.
\newblock \bibinfo{journal}{\emph{IEEE Robotics and Automation Letters}}
  \bibinfo{volume}{5}, \bibinfo{number}{4} (\bibinfo{year}{2020}),
  \bibinfo{pages}{6813--6820}.
\newblock


\bibitem[Peng et~al\mbox{.}(2022)]%
        {bokehme}
\bibfield{author}{\bibinfo{person}{Juewen Peng}, \bibinfo{person}{Zhiguo Cao},
  \bibinfo{person}{Xianrui Luo}, \bibinfo{person}{Hao Lu}, \bibinfo{person}{Ke
  Xian}, {and} \bibinfo{person}{Jianming Zhang}.}
  \bibinfo{year}{2022}\natexlab{}.
\newblock \showarticletitle{BokehMe: When Neural Rendering Meets Classical
  Rendering}. In \bibinfo{booktitle}{\emph{Proceedings of the IEEE/CVF
  Conference on Computer Vision and Pattern Recognition (CVPR)}}.
  \bibinfo{pages}{16283--16292}.
\newblock


\bibitem[Radford et~al\mbox{.}(2018)]%
        {gpt1}
\bibfield{author}{\bibinfo{person}{Alec Radford}, \bibinfo{person}{Karthik
  Narasimhan}, \bibinfo{person}{Tim Salimans}, {and} \bibinfo{person}{Ilya
  Sutskever}.} \bibinfo{year}{2018}\natexlab{}.
\newblock \showarticletitle{Improving language understanding by generative
  pre-training}.
\newblock \bibinfo{journal}{\emph{OpenAI blog}} (\bibinfo{year}{2018}).
\newblock


\bibitem[Radford et~al\mbox{.}(2019)]%
        {gpt2}
\bibfield{author}{\bibinfo{person}{Alec Radford}, \bibinfo{person}{Jeffrey Wu},
  \bibinfo{person}{Rewon Child}, \bibinfo{person}{David Luan},
  \bibinfo{person}{Dario Amodei}, \bibinfo{person}{Ilya Sutskever},
  {et~al\mbox{.}}} \bibinfo{year}{2019}\natexlab{}.
\newblock \showarticletitle{Language models are unsupervised multitask
  learners}.
\newblock \bibinfo{journal}{\emph{OpenAI blog}} (\bibinfo{year}{2019}).
\newblock


\bibitem[Ranftl et~al\mbox{.}(2021)]%
        {dpt}
\bibfield{author}{\bibinfo{person}{Ren{\'e} Ranftl}, \bibinfo{person}{Alexey
  Bochkovskiy}, {and} \bibinfo{person}{Vladlen Koltun}.}
  \bibinfo{year}{2021}\natexlab{}.
\newblock \showarticletitle{Vision transformers for dense prediction}. In
  \bibinfo{booktitle}{\emph{Proceedings of the IEEE/CVF International
  Conference on Computer Vision (ICCV)}}. \bibinfo{pages}{12179--12188}.
\newblock


\bibitem[Ranftl et~al\mbox{.}(2020)]%
        {midas}
\bibfield{author}{\bibinfo{person}{Ren{\'e} Ranftl}, \bibinfo{person}{Katrin
  Lasinger}, \bibinfo{person}{David Hafner}, \bibinfo{person}{Konrad
  Schindler}, {and} \bibinfo{person}{Vladlen Koltun}.}
  \bibinfo{year}{2020}\natexlab{}.
\newblock \showarticletitle{Towards robust monocular depth estimation: Mixing
  datasets for zero-shot cross-dataset transfer}.
\newblock \bibinfo{journal}{\emph{IEEE transactions on pattern analysis and
  machine intelligence}} \bibinfo{volume}{44}, \bibinfo{number}{03}
  (\bibinfo{year}{2020}), \bibinfo{pages}{1623--1637}.
\newblock


\bibitem[Sch\"{o}nberger and Frahm(2016)]%
        {colmapsfm}
\bibfield{author}{\bibinfo{person}{Johannes~Lutz Sch\"{o}nberger} {and}
  \bibinfo{person}{Jan-Michael Frahm}.} \bibinfo{year}{2016}\natexlab{}.
\newblock \showarticletitle{Structure-from-Motion Revisited}. In
  \bibinfo{booktitle}{\emph{Proceedings of the IEEE/CVF Conference on Computer
  Vision and Pattern Recognition (CVPR)}}. \bibinfo{pages}{4104--4113}.
\newblock


\bibitem[Sch\"{o}nberger et~al\mbox{.}(2016)]%
        {colmapmvs}
\bibfield{author}{\bibinfo{person}{Johannes~Lutz Sch\"{o}nberger},
  \bibinfo{person}{Enliang Zheng}, \bibinfo{person}{Marc Pollefeys}, {and}
  \bibinfo{person}{Jan-Michael Frahm}.} \bibinfo{year}{2016}\natexlab{}.
\newblock \showarticletitle{Pixelwise View Selection for Unstructured
  Multi-View Stereo}. In \bibinfo{booktitle}{\emph{European Conference on
  Computer Vision (ECCV)}}, Vol.~\bibinfo{volume}{9907}.
  \bibinfo{pages}{501--518}.
\newblock


\bibitem[Schuster and Paliwal(1997)]%
        {sxrnn}
\bibfield{author}{\bibinfo{person}{Mike Schuster} {and}
  \bibinfo{person}{Kuldip~K Paliwal}.} \bibinfo{year}{1997}\natexlab{}.
\newblock \showarticletitle{Bidirectional recurrent neural networks}.
\newblock \bibinfo{journal}{\emph{IEEE transactions on Signal Processing}}
  \bibinfo{volume}{45}, \bibinfo{number}{11} (\bibinfo{year}{1997}),
  \bibinfo{pages}{2673--2681}.
\newblock


\bibitem[Silberman et~al\mbox{.}(2012)]%
        {nyu}
\bibfield{author}{\bibinfo{person}{Nathan Silberman}, \bibinfo{person}{Derek
  Hoiem}, \bibinfo{person}{Pushmeet Kohli}, {and} \bibinfo{person}{Rob
  Fergus}.} \bibinfo{year}{2012}\natexlab{}.
\newblock \showarticletitle{Indoor segmentation and support inference from rgbd
  images}. In \bibinfo{booktitle}{\emph{European Conference on Computer Vision
  (ECCV)}}. Springer, \bibinfo{pages}{746--760}.
\newblock


\bibitem[Tang and Tan(2018)]%
        {banet}
\bibfield{author}{\bibinfo{person}{Chengzhou Tang} {and} \bibinfo{person}{Ping
  Tan}.} \bibinfo{year}{2018}\natexlab{}.
\newblock \showarticletitle{BA-Net: Dense Bundle Adjustment Networks}. In
  \bibinfo{booktitle}{\emph{International Conference on Learning
  Representations}}.
\newblock


\bibitem[Teed and Deng(2019)]%
        {deepv2d}
\bibfield{author}{\bibinfo{person}{Zachary Teed} {and} \bibinfo{person}{Jia
  Deng}.} \bibinfo{year}{2019}\natexlab{}.
\newblock \showarticletitle{DeepV2D: Video to Depth with Differentiable
  Structure from Motion}. In \bibinfo{booktitle}{\emph{International Conference
  on Learning Representations}}.
\newblock


\bibitem[Teed and Deng(2020)]%
        {raft}
\bibfield{author}{\bibinfo{person}{Zachary Teed} {and} \bibinfo{person}{Jia
  Deng}.} \bibinfo{year}{2020}\natexlab{}.
\newblock \showarticletitle{Raft: Recurrent all-pairs field transforms for
  optical flow}. In \bibinfo{booktitle}{\emph{European Conference on Computer
  Vision (ECCV)}}. Springer, \bibinfo{pages}{402--419}.
\newblock


\bibitem[Vaswani et~al\mbox{.}(2017)]%
        {transformer}
\bibfield{author}{\bibinfo{person}{Ashish Vaswani}, \bibinfo{person}{Noam
  Shazeer}, \bibinfo{person}{Niki Parmar}, \bibinfo{person}{Jakob Uszkoreit},
  \bibinfo{person}{Llion Jones}, \bibinfo{person}{Aidan~N Gomez},
  \bibinfo{person}{{\L}ukasz Kaiser}, {and} \bibinfo{person}{Illia
  Polosukhin}.} \bibinfo{year}{2017}\natexlab{}.
\newblock \showarticletitle{Attention is all you need}. In
  \bibinfo{booktitle}{\emph{Advances in neural information processing
  systems}}, Vol.~\bibinfo{volume}{30}.
\newblock


\bibitem[Wang et~al\mbox{.}(2021)]%
        {sfm21}
\bibfield{author}{\bibinfo{person}{Jianyuan Wang}, \bibinfo{person}{Yiran
  Zhong}, \bibinfo{person}{Yuchao Dai}, \bibinfo{person}{Stan Birchfield},
  \bibinfo{person}{Kaihao Zhang}, \bibinfo{person}{Nikolai Smolyanskiy}, {and}
  \bibinfo{person}{Hongdong Li}.} \bibinfo{year}{2021}\natexlab{}.
\newblock \showarticletitle{Deep two-view structure-from-motion revisited}. In
  \bibinfo{booktitle}{\emph{Proceedings of the IEEE/CVF Conference on Computer
  Vision and Pattern Recognition (CVPR)}}. \bibinfo{pages}{8953--8962}.
\newblock


\bibitem[Wu et~al\mbox{.}(2019)]%
        {scgan}
\bibfield{author}{\bibinfo{person}{Zhenyao Wu}, \bibinfo{person}{Xinyi Wu},
  \bibinfo{person}{Xiaoping Zhang}, \bibinfo{person}{Song Wang}, {and}
  \bibinfo{person}{Lili Ju}.} \bibinfo{year}{2019}\natexlab{}.
\newblock \showarticletitle{Spatial correspondence with generative adversarial
  network: Learning depth from monocular videos}. In
  \bibinfo{booktitle}{\emph{Proceedings of the IEEE/CVF International
  Conference on Computer Vision (ICCV)}}. \bibinfo{pages}{7494--7504}.
\newblock


\bibitem[Xian et~al\mbox{.}(2018)]%
        {kexian2018}
\bibfield{author}{\bibinfo{person}{Ke Xian}, \bibinfo{person}{Chunhua Shen},
  \bibinfo{person}{Zhiguo Cao}, \bibinfo{person}{Hao Lu}, \bibinfo{person}{Yang
  Xiao}, \bibinfo{person}{Ruibo Li}, {and} \bibinfo{person}{Zhenbo Luo}.}
  \bibinfo{year}{2018}\natexlab{}.
\newblock \showarticletitle{Monocular Relative Depth Perception With Web Stereo
  Data Supervision}. In \bibinfo{booktitle}{\emph{Proceedings of the IEEE/CVF
  Conference on Computer Vision and Pattern Recognition (CVPR)}}.
  \bibinfo{pages}{311--320}.
\newblock


\bibitem[Xian et~al\mbox{.}(2020)]%
        {kexian2020}
\bibfield{author}{\bibinfo{person}{Ke Xian}, \bibinfo{person}{Jianming Zhang},
  \bibinfo{person}{Oliver Wang}, \bibinfo{person}{Long Mai},
  \bibinfo{person}{Zhe Lin}, {and} \bibinfo{person}{Zhiguo Cao}.}
  \bibinfo{year}{2020}\natexlab{}.
\newblock \showarticletitle{Structure-Guided Ranking Loss for Single Image
  Depth Prediction}. In \bibinfo{booktitle}{\emph{Proceedings of the IEEE/CVF
  Conference on Computer Vision and Pattern Recognition (CVPR)}}.
  \bibinfo{pages}{608--617}.
\newblock


\bibitem[Xie et~al\mbox{.}(2017)]%
        {resnext}
\bibfield{author}{\bibinfo{person}{Saining Xie}, \bibinfo{person}{Ross
  Girshick}, \bibinfo{person}{Piotr Doll{\'a}r}, \bibinfo{person}{Zhuowen Tu},
  {and} \bibinfo{person}{Kaiming He}.} \bibinfo{year}{2017}\natexlab{}.
\newblock \showarticletitle{Aggregated residual transformations for deep neural
  networks}. In \bibinfo{booktitle}{\emph{Proceedings of the IEEE/CVF
  Conference on Computer Vision and Pattern Recognition (CVPR)}}.
  \bibinfo{pages}{1492--1500}.
\newblock


\bibitem[Xu et~al\mbox{.}(2018)]%
        {n13}
\bibfield{author}{\bibinfo{person}{Dan Xu}, \bibinfo{person}{Wanli Ouyang},
  \bibinfo{person}{Xiaogang Wang}, {and} \bibinfo{person}{Nicu Sebe}.}
  \bibinfo{year}{2018}\natexlab{}.
\newblock \showarticletitle{Pad-net: Multi-tasks guided
  prediction-and-distillation network for simultaneous depth estimation and
  scene parsing}. In \bibinfo{booktitle}{\emph{Proceedings of the IEEE/CVF
  Conference on Computer Vision and Pattern Recognition (CVPR)}}.
  \bibinfo{pages}{675--684}.
\newblock


\bibitem[Yin et~al\mbox{.}(2019)]%
        {vnl}
\bibfield{author}{\bibinfo{person}{Wei Yin}, \bibinfo{person}{Yifan Liu},
  \bibinfo{person}{Chunhua Shen}, {and} \bibinfo{person}{Youliang Yan}.}
  \bibinfo{year}{2019}\natexlab{}.
\newblock \showarticletitle{Enforcing geometric constraints of virtual normal
  for depth prediction}. In \bibinfo{booktitle}{\emph{Proceedings of the
  IEEE/CVF International Conference on Computer Vision (ICCV)}}.
  \bibinfo{pages}{5684--5693}.
\newblock


\bibitem[Zhang et~al\mbox{.}(2019b)]%
        {ST-CLSTM}
\bibfield{author}{\bibinfo{person}{Haokui Zhang}, \bibinfo{person}{Chunhua
  Shen}, \bibinfo{person}{Ying Li}, \bibinfo{person}{Yuanzhouhan Cao},
  \bibinfo{person}{Yu Liu}, {and} \bibinfo{person}{Youliang Yan}.}
  \bibinfo{year}{2019}\natexlab{b}.
\newblock \showarticletitle{Exploiting temporal consistency for real-time video
  depth estimation}. In \bibinfo{booktitle}{\emph{Proceedings of the IEEE/CVF
  International Conference on Computer Vision (ICCV)}}.
  \bibinfo{pages}{1725--1734}.
\newblock


\bibitem[Zhang et~al\mbox{.}(2019a)]%
        {videobokeh}
\bibfield{author}{\bibinfo{person}{Xuaner Zhang}, \bibinfo{person}{Kevin
  Matzen}, \bibinfo{person}{Vivien Nguyen}, \bibinfo{person}{Dillon Yao},
  \bibinfo{person}{You Zhang}, {and} \bibinfo{person}{Ren Ng}.}
  \bibinfo{year}{2019}\natexlab{a}.
\newblock \showarticletitle{Synthetic defocus and look-ahead autofocus for
  casual videography}.
\newblock \bibinfo{journal}{\emph{ACM Transactions on Graphics (TOG)}}
  \bibinfo{volume}{38}, \bibinfo{number}{4} (\bibinfo{year}{2019}).
\newblock


\bibitem[Zhang et~al\mbox{.}(2021)]%
        {dycvd}
\bibfield{author}{\bibinfo{person}{Zhoutong Zhang}, \bibinfo{person}{Forrester
  Cole}, \bibinfo{person}{Richard Tucker}, \bibinfo{person}{William~T Freeman},
  {and} \bibinfo{person}{Tali Dekel}.} \bibinfo{year}{2021}\natexlab{}.
\newblock \showarticletitle{Consistent depth of moving objects in video}.
\newblock \bibinfo{journal}{\emph{ACM Transactions on Graphics (TOG)}}
  \bibinfo{volume}{40}, \bibinfo{number}{4} (\bibinfo{year}{2021}),
  \bibinfo{pages}{1--12}.
\newblock


\bibitem[Zhou et~al\mbox{.}(2018)]%
        {k7}
\bibfield{author}{\bibinfo{person}{Lipu Zhou}, \bibinfo{person}{Jiamin Ye},
  \bibinfo{person}{Montiel Abello}, \bibinfo{person}{Shengze Wang}, {and}
  \bibinfo{person}{Michael Kaess}.} \bibinfo{year}{2018}\natexlab{}.
\newblock \showarticletitle{Unsupervised learning of monocular depth estimation
  with bundle adjustment, super-resolution and clip loss}.
\newblock \bibinfo{journal}{\emph{arXiv preprint arXiv:1812.03368}}
  (\bibinfo{year}{2018}).
\newblock


\bibitem[Zhu et~al\mbox{.}(2021)]%
        {newdetr}
\bibfield{author}{\bibinfo{person}{Xizhou Zhu}, \bibinfo{person}{Weijie Su},
  \bibinfo{person}{Lewei Lu}, \bibinfo{person}{Bin Li},
  \bibinfo{person}{Xiaogang Wang}, {and} \bibinfo{person}{Jifeng Dai}.}
  \bibinfo{year}{2021}\natexlab{}.
\newblock \showarticletitle{Deformable detr: Deformable transformers for
  end-to-end object detection}. In \bibinfo{booktitle}{\emph{International
  Conference on Learning Representations}}.
\newblock


\end{thebibliography}
\clearpage
\appendix
\section{Mask sampling strategies}
\label{sec:ss1}
In this section, we illustrate our ablation study on the mask sampling strategies. We adopt the random masking for training in our approach. In Table~\ref{tab:1}, we also train our \network{} in the uniform manner, which means we use fixed and uniform masking
% in training time.
during training.
For example, we will retain the fourth and eighth frames with twelve frames input. We keep the same masking ratio of $83.33\%$ for the comparison of sampling strategies.

In Table~\ref{tab:1}, we can see that our random masking strategy achieves higher depth accuracy and better temporal consistency. The random masking can be considered as a form of data argumentation. In this way, our \network{} can learn temporal correlations with various time intervals, while the uniform masking strategy can only model correlations of a fixed length of time such as $4$ frames. As a consequence, we adopt the random masking strategy for training.
\begin{table}[h]
  \caption{Ablation study on mask sampling strategies. We keep the same masking ratio of $\textbf{83.33\%}$ for the comparison of sampling strategies. In order to reduce the experimental cost, we randomly choose 40 videos for training and 10 videos for $\textbf{OPW}$ evaluation on the NYU Depth V2 dataset~\cite{nyu} in this experiment. The depth metrics are still evaluated on the public test dataset with 654 samples, which can not be compared with the results of our \network{} on the full dataset. The way of random masking achieves higher spatial accuracy and better temporal consistency. Best performance is in boldface.}
  \label{tab:1}
  \resizebox{\columnwidth}{!}{
  \begin{tabular}{lccccccc}
    \toprule
    Sampling & Rel & RMSE & $\log 10$ & $\delta_1$ & $\delta_2$ & $\delta_3$ & $OPW$ \\
    \midrule
     Uniform & $0.253$ & $\textbf{0.725}$ & $0.095$ & $0.622$ & $0.875$ & $0.965$ & $8.063$\\
     Random & $\textbf{0.221}$ & $0.738$ & $\textbf{0.093}$ & $\textbf{0.628}$ & $\textbf{0.889}$ & $\textbf{0.968}$ & $\textbf{6.965}$\\
    \bottomrule
\end{tabular}
}
\end{table}
\section{Masking ratios for inference}
\begin{figure}[htbp]
      \centering
      \centerline{\includegraphics[width=0.95\linewidth,trim=80 0 150 100,clip]{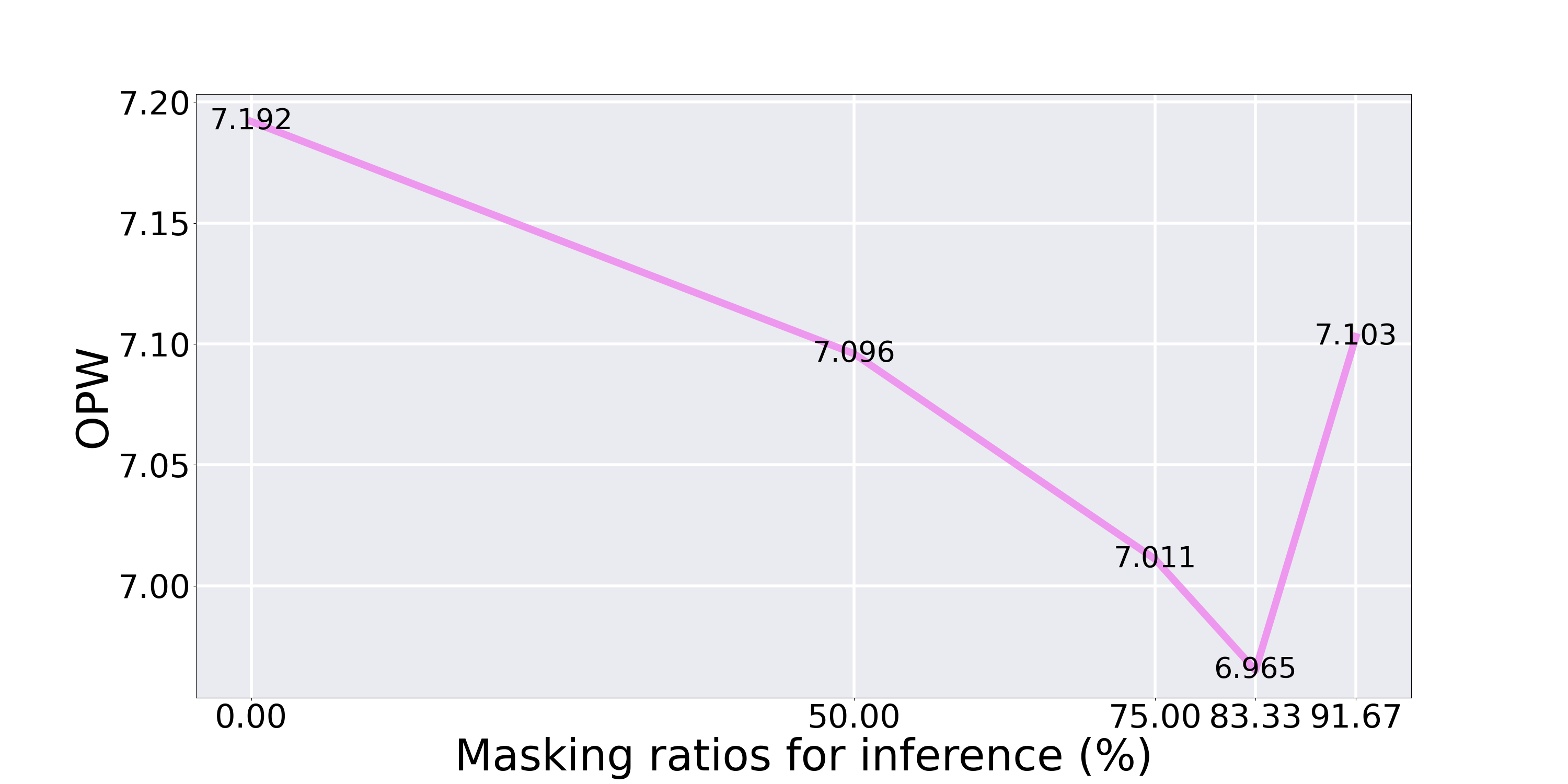}}
      \caption{Ablation study on masking ratios for inference. The X-axis
represents masking ratios and the Y-axis means OPW. Here we use the same random masking model in Table~\ref{tab:1}.}
      \label{fig:inf_rio}
\end{figure}

We also ablate our masking ratios for inference in Fig.~\ref{fig:inf_rio}. In our approach, we use uniform masking for inference to avoid randomness in our depth prediction results. For example, with twelve frames input, $83.33\%$ means that we mask ten frames and retain the fourth and eighth frames. In this experiment, we use the same model trained with $83.33\%$ random masking on 40 videos in Sec.~\ref{sec:ss1}. The $OPW$ is evaluated on the same 10 videos.

We can see that inferring with lower masking ratios causes a decrease of consistency due to higher redundancy. We also try the extreme situation: inference without masking. We directly feed input sequences without masking to our temporal structure encoder. In this way, our \network{} loses the vital mechanism of masked frames predicting. Reconstructing masked frames according to the unmasked ones plays a significant role in temporal consistency.

\begin{figure}[htbp]
    \centering
    \centerline{\includegraphics[scale=0.09,trim=30 50 30 50,clip]{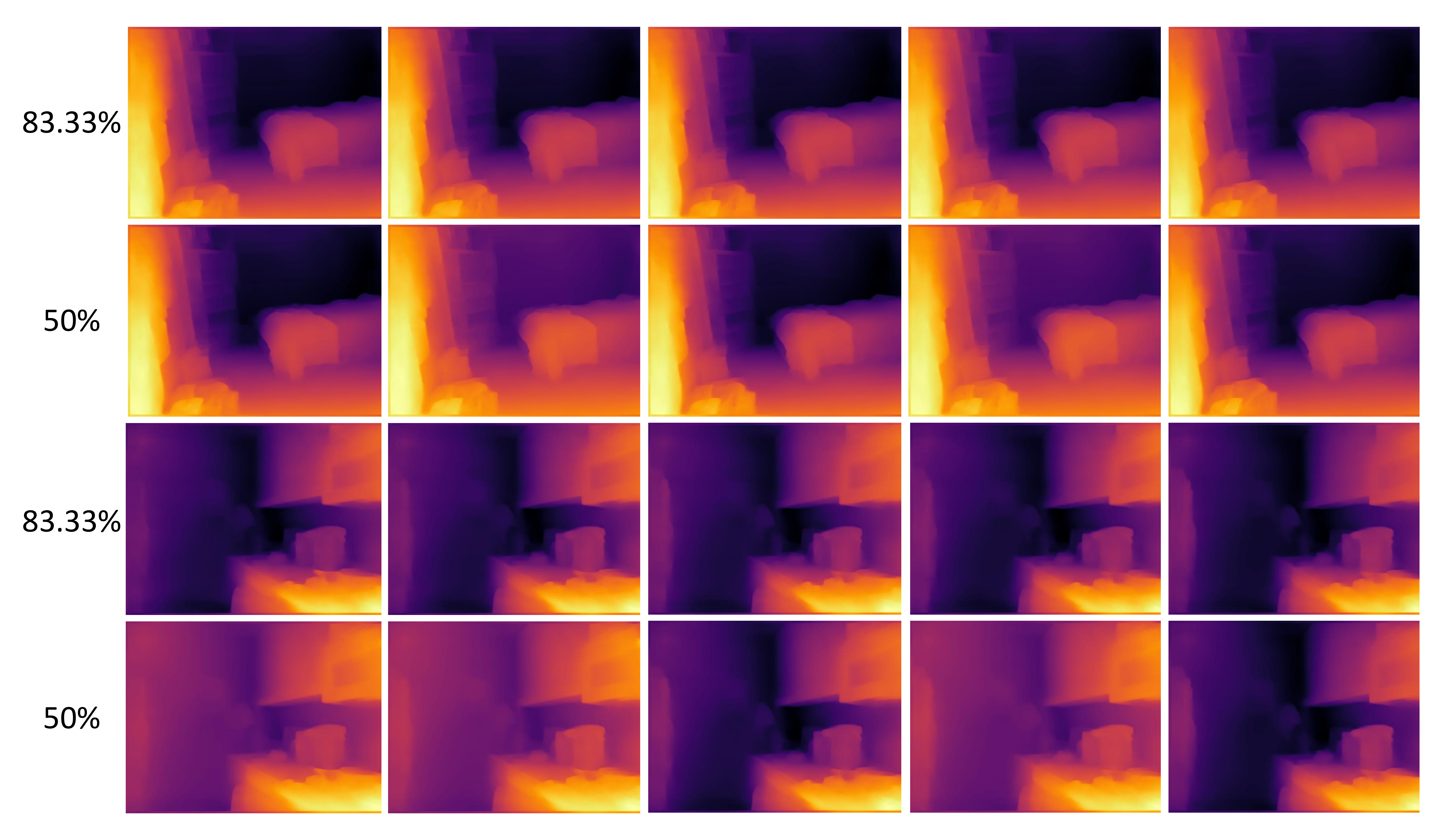}}
    \caption{Visual results comparison of different masking ratios for inference. The results are produced by our \network{} on the full NYU Depth V2 dataset~\cite{nyu}, which is trained with the random masking ratio of $\textbf{83.33\%}$.}
    \label{fig:abd}
\end{figure}

In Fig.~\ref{fig:abd}, based on our \network{} trained on the full NYU depth V2 dataset~\cite{nyu}, we compare the visual depth results of $83.33\%$ and $50\%$ masking ratios for inference. The qualitative results of $50\%$ masking ratios have worse consistency and flickering than $83.33\%$ masking ratios due to higher temporal redundancy.

\section{depth estimation metrics}
We adopt the commonly applied depth estimation metrics defined as follows:

\begin{itemize}[leftmargin=*]
\item Mean relative error (REL): $\frac{1}{n}\sum_{i=1}^n\frac{||d_i - d_i^*||_1}{d_i^*};$
\item Root mean squared error (RMSE): $\sqrt{\frac{1}{n}\sum_{i=1}^n(d_i - d_i^*)^2};$
\item Mean $\log_{10}$ error ($\log10$): $\frac{1}{n}\sum_{i=1}^n||\log_{10}d_i - \log_{10}d_i^*||_1;$
\item Accuracy with threshold $t$: Percentage of $d_i$ such that $\\max(\frac{d_i}{d_i^*},\frac{d_i^*}{d_i}) = \delta<t\in\left[1.25, 1.25^2, 1.25^3\right],$
\end{itemize}
where $n$ denotes the total number of pixels, $d_i$ and $d_i^*$ are estimated and ground truth depth of pixel $i$, respectively.
\section{Comparison with structure-from-motion methods}
We show the comparison with structure-from-motion (SFM) methods on the KITTI dataset in Table \ref{tab:sfm}. The quantitative results of structure-from-motion based methods (e.g., DeepV2D~\cite{deepv2d}) seem higher than the methods on the first four rows. However, those two kinds of methods are in different settings.

Structure-from-motion methods predict depth maps by feature matching over multiple frames. According to CVD~\cite{cvd}, this idea benefits static scenes but brings an unavoidable defect which is that these methods "do not account for dynamically moving objects". They heavily rely on explicit motion segmentation. For example, they need to mask the moving cars or people for SFM and pose estimation. When their methods are used for videos with natural scenes or obvious objects motion, those methods inevitably fail. By contrast, our method is not limited by SFM and pose estimation.%We test several videos with natural scenes or moving objects. DeepV2D completely fails and produces poor results with large depth errors.

In conclusion, SFM-based methods fit the bias of KITTI dataset, hence, previous works, such as dynamic-video-depth~\cite{dycvd} and Cao \textit{et al.}~\cite{MM21}, exclude the structure-from-motion methods in their comparison list. We just follow the same setting and add some latest works to our comparison such as CVD~\cite{cvd} and Cao \textit{et al.}~\cite{MM21}.

%Our method is not limited by optical flow and pose estimation. It can be extended to natural scenes videos. Following ST-CLSTM (ICCV 2019) and Cao \textit{et al.} (ACM MM 2021), we conduct our experiments on the NYU and KITTI datasets in our paper. It's a traditional experimental protocol in this area because there is no large scale natural scenes public video depth dataset for now. It is not a limitation of our method, but a limitation of current datasets.
\begin{table}[h]
  \caption{Comparison with structure-from-motion methods on the KITTI dataset. The structure-from-motion methods are on the last three rows and other methods are on the first four rows.}
  \label{tab:sfm}
  \resizebox{\columnwidth}{!}{
  \begin{tabular}{lccccccc}
    \toprule
    Method & Rel & RMSE & $\log 10$ & $\delta_1$ & $\delta_2$ & $\delta_3$\\
    \midrule
     ST-CLSTM~\cite{ST-CLSTM} (ICCV 2019) & $0.101$ & $4.137$ & $0.043$ & $0.890$ & $0.970$ & $0.989$\\
     CVD~\cite{cvd} (ACM SIGGRAPH 2020) & $0.130$ & $4.876$ & $-$ & $0.878$ & $0.946$ & $0.970$\\
     Cao \textit{et al.}~\cite{MM21} (ACM MM 2021)& $0.109$ & $4.366$ & $0.047$ & $0.872$ & $0.962$ & $0.986$\\
     Ours & $0.099$ & $3.832$ & $0.042$ & $0.886$ & $0.968$ & $0.989$\\
     \midrule
     BA-Net~\cite{banet} (ICLR 2018) & $0.083$ & $3.640$ & $-$ & $-$ & $-$ & $-$\\
     DeepV2D~\cite{deepv2d} (ICLR 2020) & $0.037$ & $2.005$ & $-$ & $0.977$ & $0.993$ & $0.997$\\
     Wang \textit{et al.}~\cite{sfm21} (CVPR 2021)& $0.034$ & $1.919$ & $-$ & $0.989$ & $0.998$ & $0.999$\\
    \bottomrule
\end{tabular}
}
\end{table}
\begin{table}[!h]
  \caption{Comparison with single image depth estimation methods on the KITTI dataset. The first four rows are consistent video depth methods. The last five rows are methods only for spatial depth accuracy.}
  \label{tab:ss}
  \resizebox{\columnwidth}{!}{
  \begin{tabular}{lcccccccc}
    \toprule
    Method & Rel & RMSE & $\log 10$ & $\delta_1$ & $\delta_2$ & $\delta_3$ & $OPW$ \\
    \midrule
    ST-CLSTM~\cite{ST-CLSTM} (ICCV 2019) & $0.101$ & $4.137$ & $0.043$ & $0.890$ & $0.970$ & $0.989$& $-$\\
    CVD~\cite{cvd} (ACM SIGGRAPH 2020) & $0.130$ & $4.876$ & $-$ & $0.878$ & $0.946$ & $0.970$ & $34.741$\\
    Cao \textit{et al.}~\cite{MM21} (ACM MM 2021)& $0.109$ & $4.366$ & $0.047$ & $0.872$ & $0.962$ & $0.986$ & $-$\\
    Ours & $0.099$ & $3.832$ & $0.042$ & $0.886$ & $0.968$ & $0.989$& $30.596$\\
    \midrule
    VNL~\cite{vnl} (ICCV 2019) & $0.072$ & $3.258$ & $-$ & $0.938$ & $0.990$ & $0.998$& $45.295$\\
    BTS~\cite{bts} & $0.056$ & $1.925$ & $-$ & $0.964$ & $0.994$ & $0.999$& $44.583$\\
    DPT~\cite{dpt} (ICCV 2021) & $0.062$ & $2.573$ & $-$ & $0.959$ & $0.995$ & $0.999$& $43.207$\\
    SC-GAN~\cite{scgan} (ICCV 2019) & $0.063$ & $2.129$ & $-$ & $0.961$ & $0.993$ & $0.998$& $-$\\
    AdaBins~\cite{adabins} (CVPR 2021) & $0.058$ & $2.360$ & $-$ & $0.964$ & $0.995$ & $0.999$& $43.841$\\
    \bottomrule
\end{tabular}
}
\end{table}
\section{Comparison with single image depth estimation methods}
Single image depth estimation methods~\cite{vnl,dpt,bts,midas,adabins} only take spatial depth accuracy into account and totally ignore the temporal depth consistency. As shown in Table \ref{tab:ss}, these methods achieve better performance in terms of spatial metrics, however, they suffer from obvious temporal inconsistency on video data. By contrast, consistent video depth estimation methods achieve much better temporal consistency. The core task of consistent video depth estimation is to remove flickering in video depth results. SC-GAN~\cite{scgan} seems to train their model on video data, however, their motivation and proposed solution only lie in spatial accuracy. This shows that these two types of methods are under two different settings. One is trying to achieve higher depth accuracy but totally ignoring the consistency; the other is trying to achieve consistent depth estimation of videos with good depth accuracy. In some real-world applications, e.g., 2D-to-3D video conversion~\cite{n1} and video bokeh rendering~\cite{bokehme,videobokeh}, depth consistency plays a vital role. Weird and obvious artifacts can be found if video depth is inconsistent.

Meanwhile, the training datasets and testing protocols are quite different between these two kinds of methods. For example, DPT~\cite{dpt}, which is one of the state-of-the-art models for single image depth estimation, trains on 1.4 million images. Midas~\cite{midas} is also based on mixing data from five different datasets. However, most of those datasets only contain single images. There is no such large scale public video depth dataset for now. Besides, some testing protocols are different. For example, Midas and DPT conduct scale and shift alignments for each testing image, while video depth methods such as ST-CLSTM~\cite{ST-CLSTM}, Cao \textit{et al.}~\cite{MM21}, and our methods do not.

Hence, previous works (ST-CLSTM~\cite{ST-CLSTM}, CVD~\cite{cvd}, and Cao \textit{et al.}~\cite{MM21}) exclude the single-image depth estimation methods in their comparison lists. We just follow the setting and add some latest works to our comparisons such as CVD~\cite{cvd} and Cao \textit{et al.}~\cite{MM21}.

\section{Ablation of different backbones}
We also conduct ablation study of different backbones on the KITTI dataset. The results are shown in Table~\ref{tab:bk}. Our \network{} can be easily extended to different backbones (the spatial structure feature extractor), which demonstrate the generality of our proposed method. Our \network{} achieves better performance than the model of Cao \textit{et al.}~\cite{MM21} with the same backbone.
\begin{table}[h]
  \vspace{-4pt}
  \caption{Ablation study of different backbones on the KITTI dataset. The first four rows are our methods. The last two rows are results of Cao \textit{et al.}~\cite{MM21}.}
  \vspace{-4pt}
  \label{tab:bk}
  \resizebox{\columnwidth}{!}{
  \begin{tabular}{lcccccccc}
    \toprule
    Method & Backbone & Rel & RMSE & $\log 10$ & $\delta_1$ & $\delta_2$ & $\delta_3$\\
    \midrule
     Ours & ResNet18 & $0.105$ & $3.936$ & $0.045$ & $0.875$ & $0.965$ & $0.988$\\
     Ours & ResNet50 & $0.105$ & $3.893$ & $0.044$ & $0.876$ & $0.965$ & $0.988$\\
     Ours & ResNet101 & $0.101$ & $3.868$ & $0.043$ & $0.882$ & $0.967$ & $0.989$\\
     Ours & ResNext101 & $0.099$ & $3.828$ & $0.042$ & $0.886$ & $0.968$ & $0.989$\\
     \midrule
     Cao \textit{et al.}~\cite{MM21} (ACM MM 2021) & ResNet18 & $0.109$ & $4.366$ & $0.047$ & $0.872$ & $0.962$ & $0.986$\\
     Cao \textit{et al.}~\cite{MM21} (ACM MM 2021) & ResNet101 & $0.106$ & $4.243$ & $0.045$ & $0.879$ & $0.964$ & $0.986$\\
    \bottomrule
    \vspace{-10pt}
\end{tabular}
}
\end{table}
\vspace{-10pt}
\section{Qualitative depth results}
\begin{figure*}[htbp]
    \centering
    \includegraphics[scale=0.130,trim=35 0 0 0,clip]{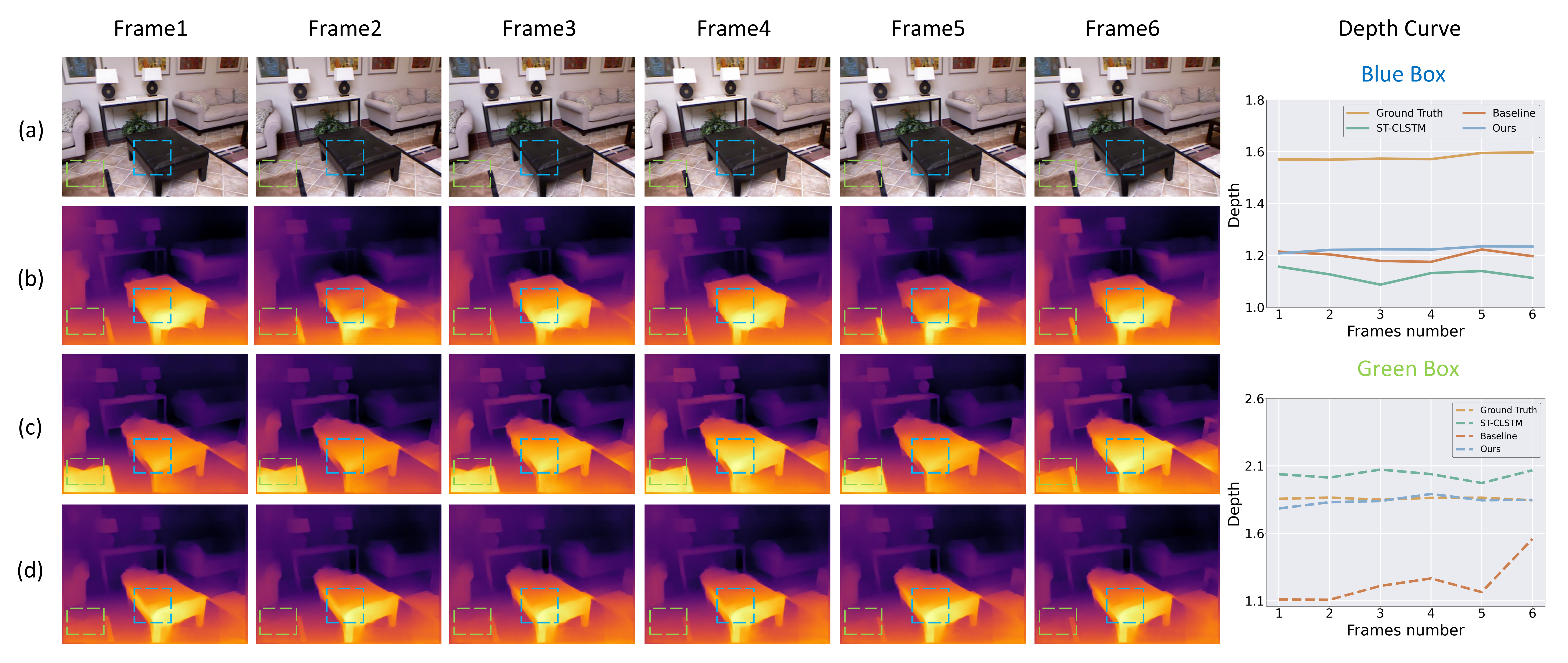}
\caption{Qualitative depth results on the NYU Depth V2 dataset~\cite{nyu}. The four rows are: (a) RGB inputs; (b) Results of ST-CLSTM~\cite{ST-CLSTM}; (c) Baseline results; (d) Results of our FMNet. We highlight regions with obvious difference in dashed rectangular. For better comparison, we draw depth curves on the last column. Each curve represents depth value for the center point of a certain box in the input frames.}
\label{fig:n1}
\end{figure*}

\begin{figure*}[htbp]
    \centering
    \includegraphics[scale=0.130,trim=35 0 0 0,clip]{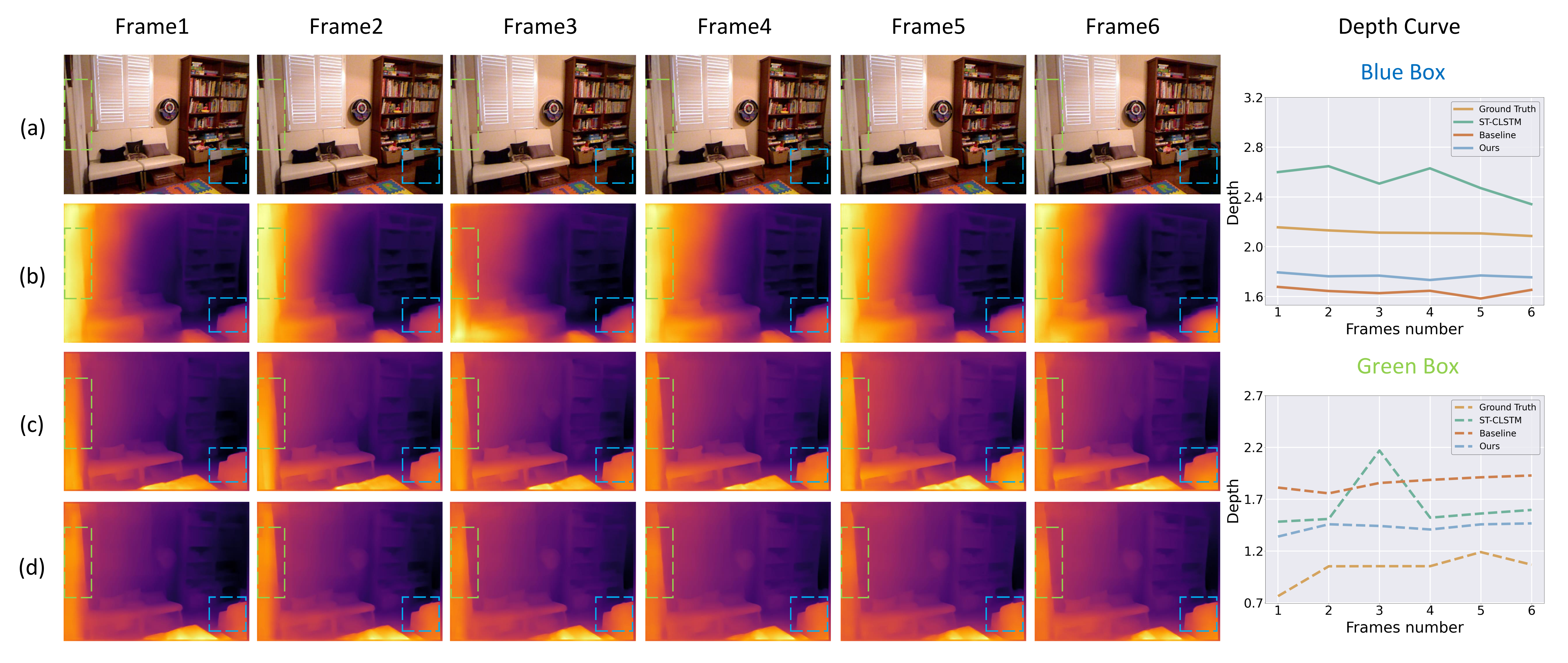}
\caption{Qualitative depth results on the NYU Depth V2 dataset~\cite{nyu}. The four rows are: (a) RGB inputs; (b) Results of ST-CLSTM~\cite{ST-CLSTM}; (c) Baseline results; (d) Results of our FMNet. We highlight regions with obvious difference in dashed rectangular. For better comparison, we draw depth curves on the last column. Each curve represents depth value for the center point of a certain box in the input frames.}
\label{fig:n2}
\end{figure*}

\begin{figure*}[htbp]
    \centering
    \includegraphics[scale=0.130,trim=35 0 0 0,clip]{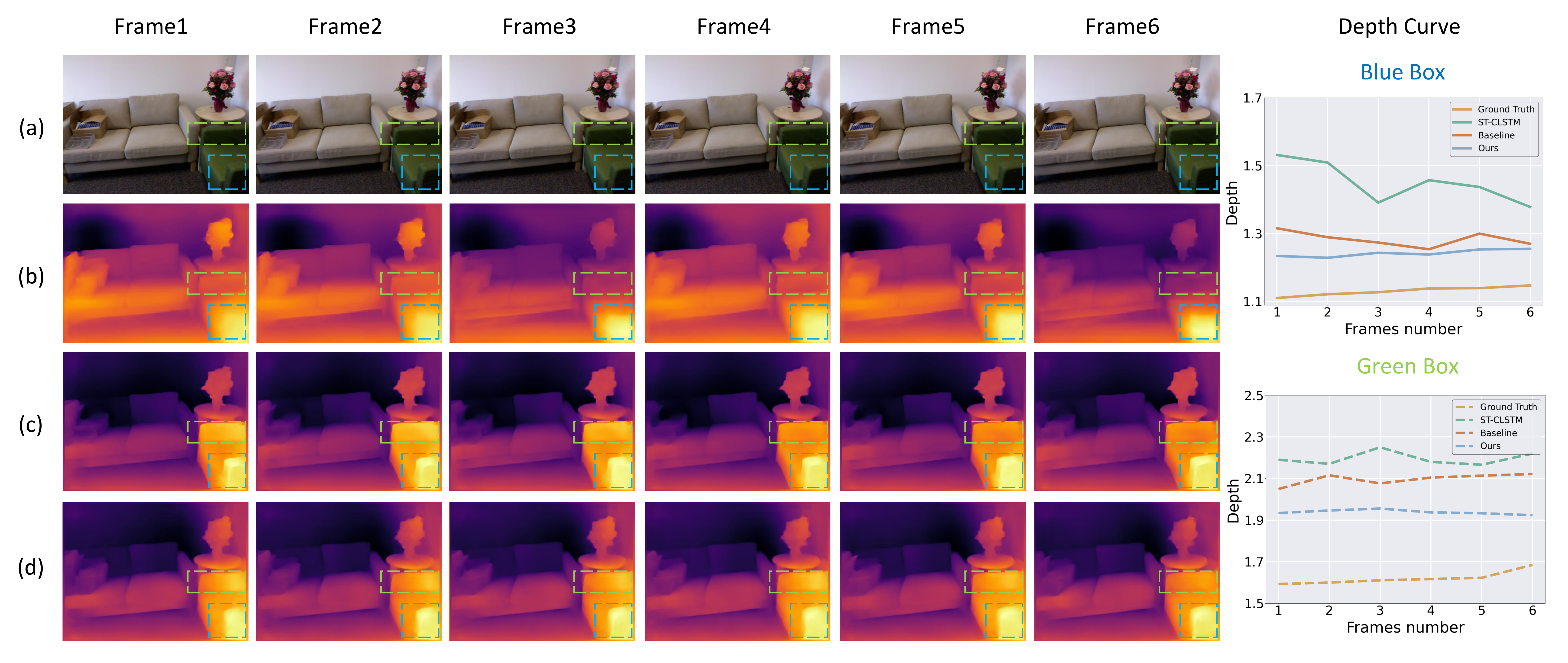}
\caption{Qualitative depth results on the NYU Depth V2 dataset~\cite{nyu}. The four rows are: (a) RGB inputs; (b) Results of ST-CLSTM~\cite{ST-CLSTM}; (c) Baseline results; (d) Results of our FMNet. We highlight regions with obvious difference in dashed rectangular. For better comparison, we draw depth curves on the last column. Each curve represents depth value for the center point of a certain box in the input frames.}
\label{fig:n3}
\end{figure*}

\begin{figure*}
    \centering
    \includegraphics[scale=0.130,trim=30 60 0 0,clip]{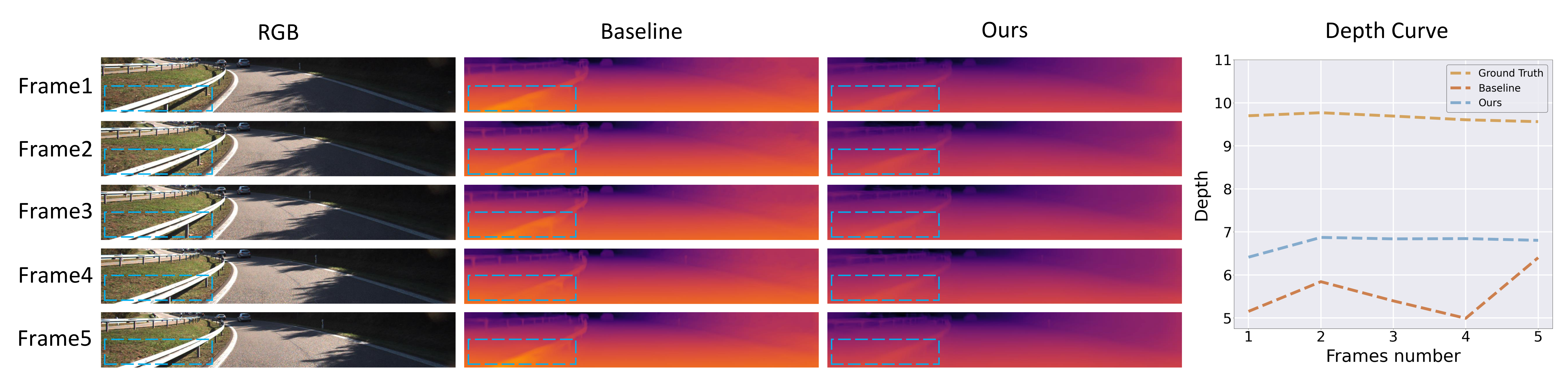}
\caption{Qualitative depth results on the KITTI dataset~\cite{kitti}. We highlight regions with obvious difference in dashed rectangular. For better comparison, we draw depth curves on the last column. Each curve represents depth value for the center point of a certain box in the input frames.}
\label{fig:k1}
\end{figure*}

\begin{figure*}
    \centering
    \includegraphics[scale=0.130,trim=30 0 0 0,clip]{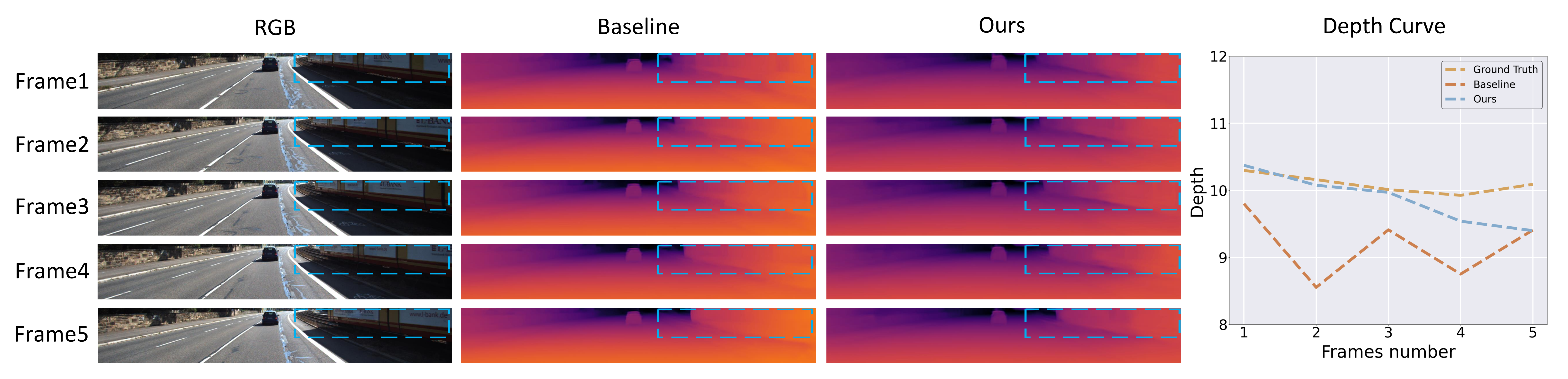}
\caption{Qualitative depth results on the KITTI dataset~\cite{kitti}. We highlight regions with obvious difference in dashed rectangular. For better comparison, we draw depth curves on the last column. Each curve represents depth value for the center point of a certain box in the input frames.}
\label{fig:k2}
\end{figure*}

\begin{figure*}
    \centering
    \includegraphics[scale=0.130,trim=30 0 0 0,clip]{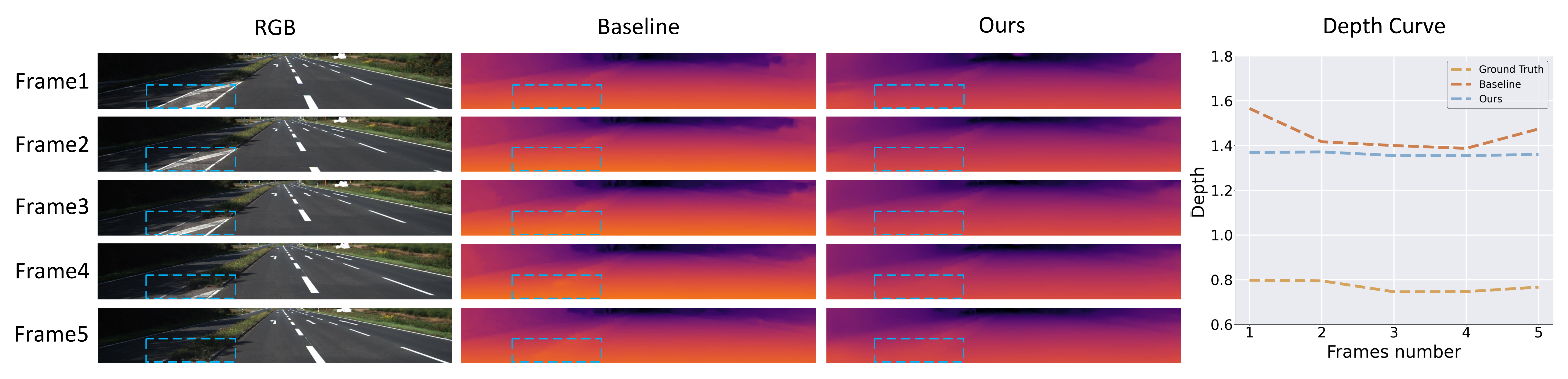}
\caption{Qualitative depth results on the KITTI dataset~\cite{kitti}. We highlight regions with obvious difference in dashed rectangular. For better comparison, we draw depth curves on the last column. Each curve represents depth value for the center point of a certain box in the input frames.}
\label{fig:k3}
\end{figure*}

We show additional qualitative depth results on the NYU Depth V2 dataset~\cite{nyu} in Fig.~\ref{fig:n1}, Fig.~\ref{fig:n2}, and Fig.~\ref{fig:n3}. Visual results on the KITTI dataset~\cite{kitti} are in Fig.~\ref{fig:k1}, Fig.~\ref{fig:k2}, and Fig.~\ref{fig:k3}. We compare the results of ST-CLSTM~\cite{ST-CLSTM}, our baseline, and our \network{}. We highlight regions with obvious difference in dashed rectangular. For better comparison, we draw depth curves on the last column. Our FMNet shows higher spatial accuracy and better temporal consistency.
\section{Depth predictor}
\begin{figure*}[htbp]
    \centering
    \includegraphics[scale=0.50,trim=30 10 30 30,clip]{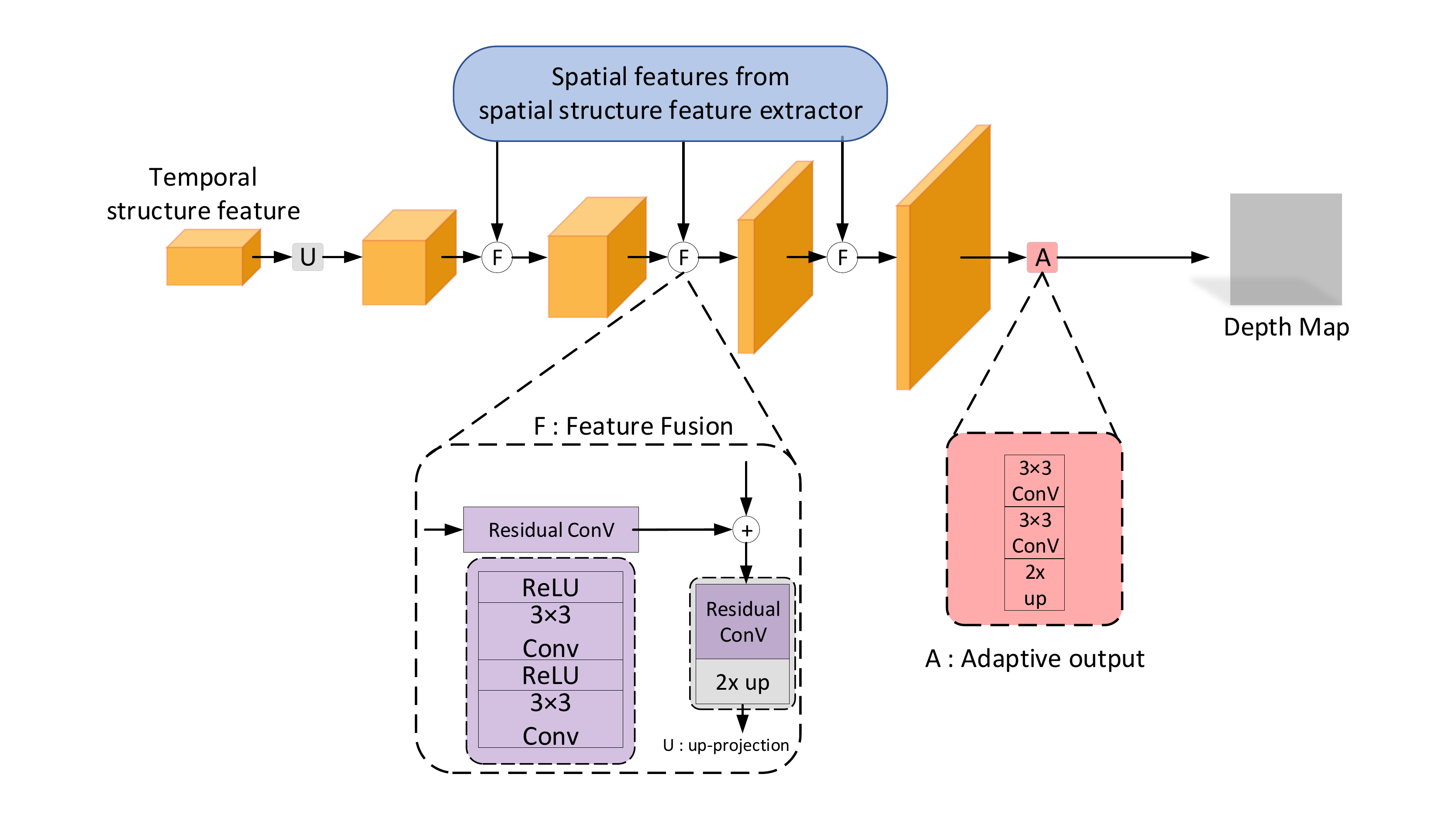}
\caption{Depth predictor.}
\label{fig:dp}
\end{figure*}

The architecture of our depth predictor is illustrated in Fig.~\ref{fig:dp}. To fuse the spatial and temporal structure features, we use the feature fusion module (FFM)~\cite{FFM1,FFM2} and skip connection from the spatial structure feature extractor to the depth predictor. The temporal structure features could improve the inter-frames temporal consistency and the spatial features could help to reconstruct the detailed information in our depth results. The adaptive output module adjusts the channel numbers and restores the depth results.



\end{document}